\PassOptionsToPackage{table,usenames,dvipsnames}{xcolor}
\documentclass[sigconf]{acmart}

\copyrightyear{2025}
\acmYear{2025}
\setcopyright{cc}
\setcctype{by}
\acmConference[KDD '25]{Proceedings of the 31st ACM SIGKDD Conference on Knowledge Discovery and Data Mining V.2}{August 3--7, 2025}{Toronto, ON, Canada}
\acmBooktitle{Proceedings of the 31st ACM SIGKDD Conference on Knowledge Discovery and Data Mining V.2 (KDD '25), August 3--7, 2025, Toronto, ON, Canada}
\acmDOI{10.1145/3711896.3737380}
\acmISBN{979-8-4007-1454-2/2025/08}

\AtBeginDocument{%
  }
\usepackage{subcaption}
\usepackage{makecell}
\usepackage{multirow}
\usepackage{float}
\usepackage{balance}
\usepackage{enumitem} 
\setlist[itemize]{leftmargin=7pt,labelsep=3pt,noitemsep,topsep=5pt}
\setlist[enumerate]{leftmargin=20pt,labelsep=10pt,noitemsep,topsep=5pt}

\begin{document}


\title{IrrMap: A Large-Scale Comprehensive Dataset for Irrigation Method Mapping }


\author{Nibir Chandra Mandal}
\authornote{Both authors contributed equally to this research.}
\orcid{0009-0008-9119-7043}
\author{Oishee Bintey Hoque}
\authornotemark[1]
\authornote{Corresponding Author}
\orcid{0000-0002-4646-3217}
\email{wyr6fx@virginia.edu}
\email{gza5rdr@virginia.edu}
\affiliation{%
  \institution{Dept. of Computer Science, \\University of Virginia}
  \city{Charlottesville}
  \state{VA}
  \country{USA}
}

\author{Abhijin Adiga}
\orcid{0000-0002-9770-034X}
\email{abhijin@virginia.edu}
\affiliation{%
  \institution{Biocomplexity Institute,\\ University of Virginia}
  \city{Charlottesville}
  \state{VA}
  \country{USA}
}

\author{Samarth Swarup}
\orcid{0000-0003-3615-1663}
\author{Mandy L. Wilson}
\orcid{0000-0002-4778-5744}
\email{swarup@virginia.edu}
\email{alw4ey@virginia.edu}
\affiliation{%
  \institution{Biocomplexity Institute,\\ University of Virginia}
  \city{Charlottesville}
  \state{VA}
  \country{USA}
}


\author{Lu Feng}
\orcid{0000-0002-4651-8441}
\author{Yangfeng Ji}
\orcid{0000-0002-7793-486X0000-0002-4651-8441}
\email{lu.feng@virginia.edu}
\email{yj3fs@virginia.edu}
\affiliation{%
  \institution{Dept. of Computer Science,\\University of Virginia}
  \city{Charlottesville}
  \state{VA}
  \country{USA}
  }


\author{Miaomiao Zhang}
\email{mz8rr@virginia.edu}
\orcid{0000-0003-0457-3335}
\affiliation{%
  \institution{Dept. of Computer Science,\\Dept. of Electrical and Computer Engineering, \\ University of Virginia}
  \city{Charlottesville}
  \state{VA}
  \country{USA}
  }

\author{Geoffrey Fox}
\orcid{0000-0003-1017-1391}
\author{Madhav Marathe}
\orcid{0000-0003-1653-0658}
\email{vxj6mb@virginia.edu}
\email{marathe@virginia.edu}
\affiliation{%
  \institution{Dept. of Computer Science, \\Biocomplexity Institute,\\ University of Virginia}
  \city{Charlottesville}
  \state{VA}
  \country{USA}
  }


\renewcommand{\shortauthors}{Mandal et al.}

\begin{abstract}
We introduce IrrMap, the first large-scale dataset~(1.1 million patches)
for irrigation method mapping across regions. IrrMap consists of multi-resolution satellite imagery from LandSat and Sentinel, along with key auxiliary data such as crop type, land use, and vegetation indices. 
The dataset spans 1,687,899 farms and 14,117,330 acres across multiple western U.S. states from 2013 to 2023, providing a rich 
and diverse foundation for irrigation analysis and ensuring geospatial alignment and quality control. 
The dataset is ML-ready, with standardized 224×224 GeoTIFF patches, the multiple input data layers, carefully chosen 
train-test-split data, and accompanying dataloaders for seamless deep learning model training and
benchmarking in irrigation mapping. The dataset is also accompanied by a complete pipeline for dataset generation, enabling 
researchers to extend IrrMap to new regions for irrigation data collection or adapt it with minimal effort for other similar 
applications in agricultural and geospatial analysis. We also analyze the irrigation method distribution across crop groups, 
spatial irrigation patterns (using Shannon diversity indices), and irrigated area variations for both LandSat and Sentinel, 
providing insights into regional and resolution-based differences. To promote further exploration, we openly release IrrMap, 
along with the derived datasets, benchmark models, and pipeline code, through a GitHub repository: \textcolor{blue}{\textit{\url{https://github.com/Nibir088/IrrMap}}} and 
Data repository: \textcolor{blue}{\textit{\url{https://huggingface.co/Nibir/IrrMap}}}, providing comprehensive documentation and implementation details.
\end{abstract}





\maketitle
\section{Introduction}
\begin{figure}
    \centering
    \includegraphics[width=.5\textwidth]{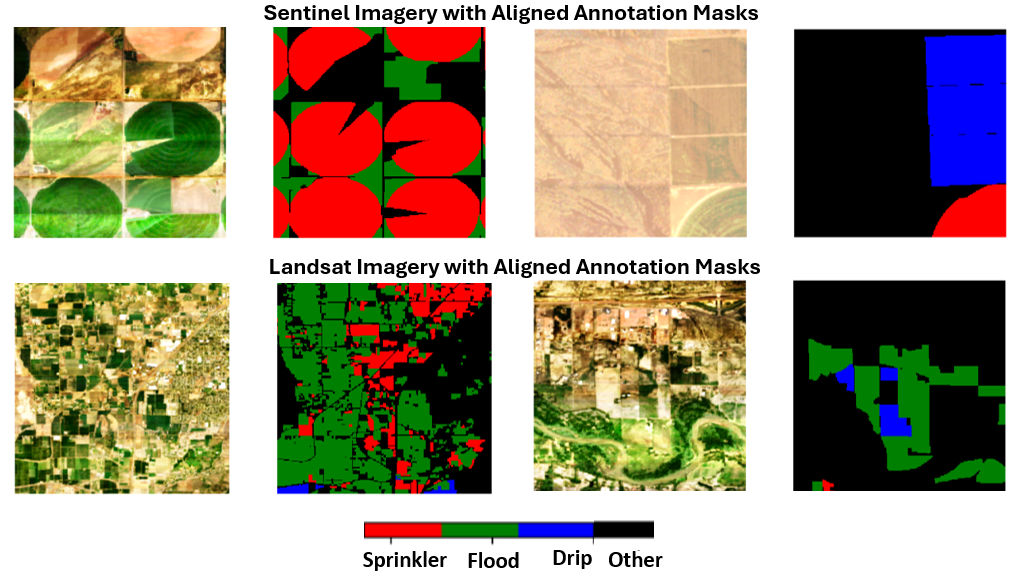}
    \caption{Spatial diversity and structural heterogeneity of irrigation: Multiple irrigation types can 
    co-occur in small areas along with non-irrigated areas. Even among fields
    with the same type, we see diversity in the way irrigation is implemented~(for 
    e.g., central
    pivot vs. non-central pivot). The spatial resolution of the imagery sources
    can adequately represent small fields as well (see 2nd row). 
    }
    \label{fig:irr_type}
\end{figure}

\begin{figure*}[!ht]
    \centering
    \includegraphics[width=0.84\linewidth]{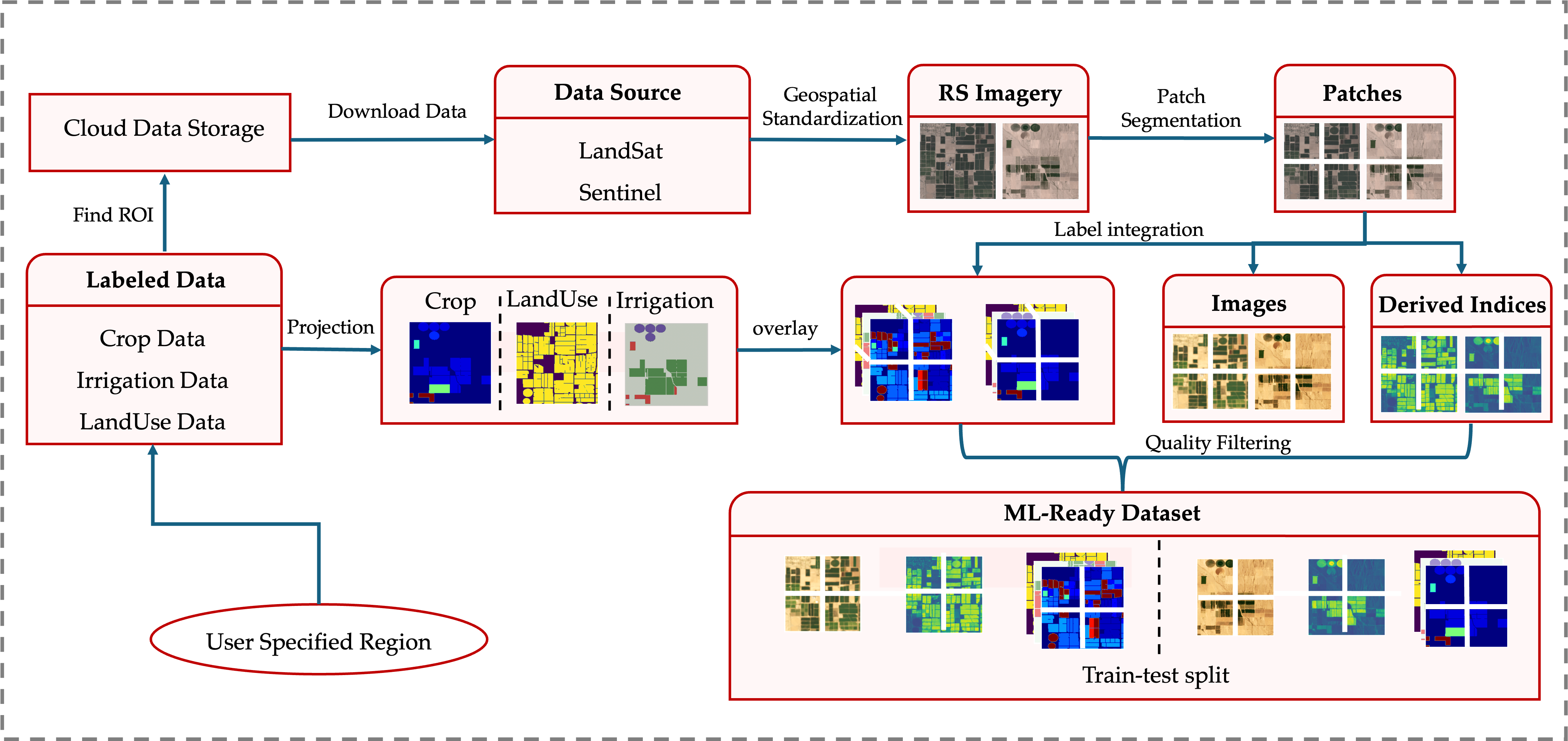}
    \caption{Pipeline for IrrMap dataset acquisition and preparation. 
    }
    \label{fig:irrMap}
\end{figure*}


Global freshwater demand has increased six-fold over the past century and
continues to increase at a rate of 1\% per year ~\cite{water2020united,ritchie2018water}. 
Irrigation accounts for 70\% of freshwater withdrawals and 90\% of consumptive 
water use worldwide ~\cite{foley2011solutions,zhang2022estimation,siebert2010groundwater}. One of 
the main factors influencing water availability in irrigated areas is the method 
of on-farm irrigation used~\cite{leng2017significant,Ippolito}. For instance,
widely used surface irrigation technologies (such as flood or furrow irrigation)
pose challenges related to irrigation efficiency and downstream water quality,
whereas sprinkler systems are often more efficient and less harmful to downstream
water quality but require high initial investments, limiting their widespread 
adoption. Given that irrigated farmland is declining in the western United
States as surface water supplies dry up and competition for water from urban,
industrial, and environmental sectors intensifies ~\cite{maupin2018summary,engelbert2023water}, accurate irrigation method identification is crucial for monitoring irrigation patterns and sustainable water resource allocation, particularly in the western United States.

Deep learning~(DL) with remote sensing has shown significant potential
in various geospatial agricultural mapping tasks~\cite{jin2017mapping,weiss2020remote}, such as crop classification~\cite{mortensen2016semantic, milioto2017real,di2017automatic},
farm boundary detection, farm infrastructure detection, and 
species mapping~\cite{hung2014feature,adiga2023robust,jia2019recurrent,kussul2017deep,handan2019deep,bueno2023mapping}. 
However,  its application in irrigation method classification remains limited due to the absence 
of a large-scale ML-ready irrigation method dataset. Existing datasets like LANID and 
IrrMapper provide 30-meter resolution maps but only distinguish between irrigated and 
non-irrigated lands without differentiating 
irrigation methods~\cite{ketchum2020irrmapper,xie2021LandSat}. 

Creating a large-scale irrigation dataset across different regions is particularly
challenging. Firstly, there is no centralized database; data for different states in the US
can potentially come from different sources. Therefore, we see variations in classification
systems, spatial resolutions, and  labeling standards of different agencies (See 
Figure~\ref{fig:irr_type}). For instance, 
what is classified as ``Center Pivot'' irrigation in Arizona may be labeled as ``Big Gun''
irrigation in Washington, which requires a systematic approach to ensure consistent labeling 
across the region. Moreover, significant variation in farm sizes poses a challenge 
in balancing high-resolution mapping for small farms while ensuring comprehensive 
regional coverage. Textural patterns of irrigation type could be heavily
influenced by crop type, soil, etc. Additional contextual information, such as crop types and vegetation indices, can be beneficial to improve classification accuracy~\cite{geographyrealm_ndwi,utah2021drip}.


\subsection{Our Contributions}
In this work, we introduce IrrMap, a large-scale comprehensive dataset for irrigation mapping 
designed to ($i$)~support the development of machine learning~(ML) models that identify various irrigation methods, and 
($ii$)~accelerate mapping various agricultural infrastructure to support resilience in food systems
and water availability.
Our dataset includes satellite imagery along with key variables relevant to irrigation mapping, such as crop type, 
land use data, and derived vegetation indices. It spans multiple states across the U.S., covering
1,668,899 farms and 11,443,492 acres from 2013 to 2023. To the best of our knowledge, IrrMap is the 
first and largest dataset dedicated to data-driven irrigation method mapping. As discussed below,
our work addresses each of the criteria laid out in the call for paper, namely: 
($i$) \emph{Accessibility}: the dataset is public (\href{https://huggingface.co/Nibir/IrrMap/tree/main}{\textcolor{blue}{Huggingface}}); 
($ii$) \emph{Quality and Documentation}: Automated and manual quality checks were conducted, along 
with documentation of the dataset, pipeline and analysis; 
($iii$) \emph{Impact}: The multi-source dataset and derived models can be extended for national-scale mapping of irrigation and related problems as well
as promote methodological advances (outlined in Section~\ref{sec:app});
($iv$) \emph{Ethics and Fairness}:  Our collected data is based solely on publicly available sources and is provided for research purposes, ensuring transparency and reproducibility

\noindent
\textbf{Multi-resolution satellite imagery.}
To accommodate diverse research needs, we integrate multi-resolution data from 
LandSat~(30m × 30m) and Sentinel~(10m × 10m), ensuring a balance between detail 
and scalability. High-resolution imagery enables fine-grained analysis of irrigation 
patterns and small-scale variations, while lower-resolution data facilitates 
large-scale assessments with reduced computational overhead, making it ideal for regional and
national studies. Besides, LandSat data is available for more years compared to Sentinel data
allowing for more detailed temporal evolution studies of irrigation. While LandSat and Sentinel are at coarser spatial resolution, the spatial and
temporal coverages of these sources is much higher than the publicly avaliable higher 
resolution images.

\noindent
\textbf{Irrigation method annotation.} Since irrigation label data is not centralized, 
we manually collected mapping data for four states— Arizona (AZ), Colorado (CO), 
Washington (WA), and Utah~(UT)— along with land use and crop layer datasets. These 
datasets were carefully cleaned and integrated with the imagery. 

\noindent
\textbf{Agricultural related indices and masks.}
Additionally, we computed twelve vegetation indices to further enhance the irrigation mapping 
task. By providing these indices, IrrMap ensures accessibility to researchers beyond remote 
sensing experts, making it easier for hydrologists, agronomists, and ML practitioners to 
leverage the dataset without requiring additional domain-specific preprocessing. 

\noindent
\textbf{ML-ready data processing.} 
Beyond providing data, IrrMap also includes a complete pipeline for generating ML-ready 
datasets in new regions, eliminating manual efforts for researchers, particularly for irrigation 
mapping, and with minor modifications for similar applications. 
Following our pipeline~(Figure~\ref{fig:irrMap}), 
given a Region of Interest (ROI), LandSat and Sentinel images are downloaded from USGS 
Earth Explorer\footnote{https://earthexplorer.usgs.gov/}. We then process a total of 25.6 TB of large image tiles into high-quality one 
million~224×224 GeoTIFF patches (~6.2 TB), ensuring precise geospatial alignment  and rigorous 
quality control, before combining them with their corresponding labels. 

\noindent
\textbf{Data analysis.}
To better understand how crop type influences irrigation type mapping and how irrigation method representation varies across different dataset resolutions, we conduct a preliminary analysis of irrigation method distribution across crop groups and spatial patterns, examining irrigation preferences across geographic regions and variations in irrigation diversity across datasets with different spatial resolutions. 

\noindent
\textbf{Benchmarking.}
We also establish benchmarks using deep learning~(DL) models trained on different input 
layers, including RGB, RGB with Crop Type, RGB with Land Use, and RGB with NDVI, 
demonstrating their effectiveness in irrigation mapping tasks. 


\noindent
\textbf{Data availability.}
To encourage further research and facilitate the development of similar datasets, we openly publish IrrMap, including dataloaders, trained models, benchmark results, and the complete pipeline for dataset generation and training the benchmark models. We also provide a \textcolor{blue}{\textit{\url{https://github.com/Nibir088/IrrMap}}} and 
Data repository: \textcolor{blue}{\textit{\url{https://huggingface.co/Nibir/IrrMap}}} with comprehensive documentation on dataset usage and model implementation. These resources serve as a valuable reference for expanding irrigation mapping efforts across the globe.

\begin{figure*}[!h]
    \centering
     \begin{subfigure}{.58\textwidth}  
    \small
        \begin{tabular}{lccccclc}
            \toprule
            \textbf{State} & \textbf{Year Range} & \multicolumn{4}{c}{\textbf{Irrigation Type (\%)}} & \textbf{Area (acres)} & \textbf{$\#$Farms} \\
            \cmidrule(r){3-6}
            & & Drip & Sprinkler & Flood & Other &  \\
            \midrule
            AZ    & 2013-17  & 7.84  & 20.05  & 45.67  & 26.43  & 566,340.14   & 16,991  \\
            CO    & 2018-20  & 0.15  & 40.24  & 58.59  & 1.02   & 2,560,487.77  & 131,585 \\
            UT    & 2021-23  & 0.08  & 34.22  & 30.60  & 35.10  & 2,673,838     & 794,202 \\
            WA    & 2015-20  & 1.70  & 20.41  & 1.93   & 75.96  & 8,316,664.40  & 726,121 \\
            \midrule
            \textbf{Total} & - & 1.36  & 26.60  & 19.39 & 52.65 & 14,117,330.31 & 1,687,899 \\
            \bottomrule
        \end{tabular}
        \caption{}
        \label{tab:summary_data}
        \end{subfigure}
        ~~~
        \begin{subfigure}{.35\textwidth}
         \centering
         \includegraphics[width=\textwidth]{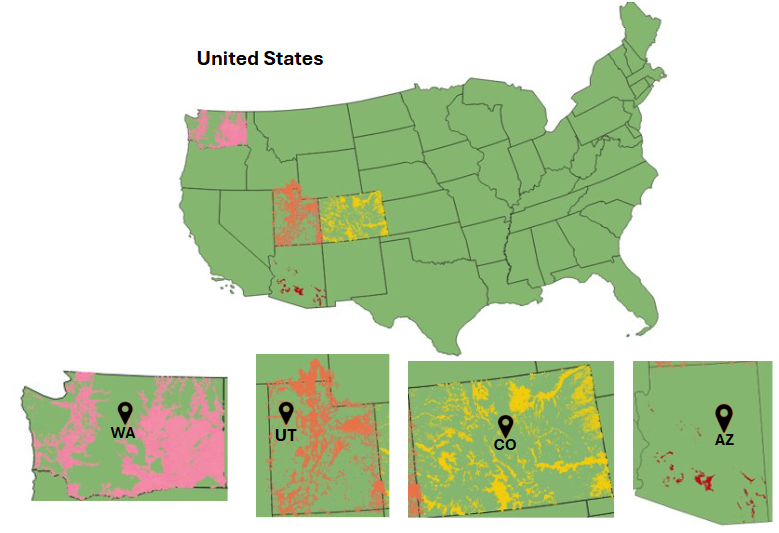} 
         \caption{}
         \label{fig:coverage_area}
        \end{subfigure}
    \caption{Spatial and temporal distribution of irrigation dataset. (a) Statistical summary of irrigation type coverage by state, showing percentage distribution across irrigation types, total irrigated area, and farm count. Note the significant class imbalance across irrigation categories. (b) Geographic distribution of study areas within the United States.}
\end{figure*}

\begin{table*}[h]
 \small
    \centering
    \renewcommand{\arraystretch}{1.2}
     \caption{Overview of the Irrigation Mapping Dataset, including spectral image information from LandSat and Sentinel satellites, along with auxiliary data such as derived indices, irrigation mapping, crop data layers, and land use information.}
    \begin{tabular}{|r|p{12cm}|l|}
   
        \toprule
        \textbf{Data Category} & \textbf{Description} & \textbf{Patch Shape} \\
        \midrule
        \multirow{4}{*}{\centering Spectral Image} & LandSat 30m: Coastal (Band 1), Blue (Band 2), Green (Band 3), Red (Band 4), NIR (Band 5), SWIR-1 (Band 6), SWIR-2 (Band 7), Thermal (Band 10 \& 11) & \multirow{2}{*}{224x224x9} \\
        \cline{2-3}
        & Sentinel 10m: Coastal (Band 1), Blue (Band 2), Green (Band 3), Red (Band 4), Red Edge 1 (Band 5), Red Edge 2 (Band 6), Red Edge 3 (Band 7), NIR (Band 8), SWIR-1 (Band 11), SWIR-2 (Band 12) & \multirow{2}{*}{224x224x10} \\
        \midrule
        \multirow{7}{*}{\centering Auxiliary Information} & Irrigation Map: Other (0), Flood (1), Sprinkler (2), Drip (3) & \multirow{1}{*}{224x224x4} \\
        \cline{2-3}
        & Derived Indices: NDVI, EVI, GNDVI, SAVI, MSAVI, RVI, CIgreen, NDWI, PRI, OSAVI, WDRVI, NDTI & \multirow{1}{*}{224x224x12} \\ 
        \cline{2-3}
        & Crop Data: Alalfa, Cereals, Cover Crop, Fibres, Fruits, Grass, Green House, Herb Group, Horticulture, Nursery, Nuts, Oil-bearing Crops, Orchard, Pulses, Roots and Tubers, Shrub Plants, Sugar Crops, Vegetables, Vinetard, Unknown. (One-Hot Encoded, Mapped with Each Patch) & \multirow{3}{*}{224x224x21} \\
        \cline{2-3}
        & Land Use: Binary Mask & 224x224x1 \\ 
        \bottomrule
    \end{tabular}
   
    \label{tab:dataset_overview}
\end{table*}
\section{Related Work}

Current irrigation mapping products provide spatial data without detailed irrigation 
methods~\cite{salmon2015global, siebert2015globaldata}. While remote sensing has been utilized 
to map irrigated fields, especially in areas of mixed agriculture~\cite{bazzi2019mapping}, 
distinguishing between irrigation types remains challenging due to landscape complexity and 
subtle practice variations. In addition, existing datasets for irrigation has offered large 
spatial coverage but lacks high-resolution imagery or often suffer from outdated 
data~\cite{meier2018global,bontemps2015multi}. Global Irrigated Area 
Map~(GIAM)~\cite{doll2000digital,thenkabail2009global}, Global Map of Irrigation 
Areas~(GMIA)~\cite{siebert2013update}, Global Land Cover 
Characteristics~(GLCC)~\cite{loveland2000development}, and Global Food-Support Analysis Data (GFSAD) have global spatial coverage for irrigated lands but use one-kilometer resolution \cite{teluguntla2015global}. MIrAD (250-meter resolution) relies on census-based estimates, leading to classification error \cite{brown2014merging,pervez2010mapping}. Although AIM-HPA (30-meter resolution) provides high-resolution irrigation data, its coverage is limited to the High Plains \cite{deines2019mapping}. Recent work on LandSat-based irrigation datasets, LANID (30-meter resolution) and IrrMapper, contains irrigated lands without labels for irrigation methods \cite{xie2021LandSat,ketchum2020irrmapper}. 

Deep learning and remote sensing based mapping has been applied recently for 
various agricultural applications, leveraging satellite and aerial imagery. Some
active areas include crop-type
and crop-land detection~\cite{jia2019recurrent,mortensen2016semantic,campos2020understanding}. 
In the context of livestock agriculture, there have 
been some works in mapping large farms due to environmental concerns~\cite{bueno2023mapping} and mixed 
farming~\cite{handan2019deep}. Farm boundary detection uses 
segmentation techniques applied to satellite imagery, leveraging edge detection and 
deep learning models to delineate agricultural 
plots~\cite{farm_boundary,field_extraction}. Mapping invasive species is another emerging 
area~\cite{adiga2023robust,di2017automatic,kussul2017deep,milioto2017real}. 
Although these datasets provide extensive agricultural monitoring capabilities, 
large-scale irrigation mapping data remains largely unavailable. 
The absence of such data hinders efforts to accurately assess water usage, monitor 
drought impacts, and optimize irrigation efficiency. Given the increasing concerns over water scarcity and sustainable agriculture, comprehensive irrigation mapping would be critical for policymakers, water resource managers, and farmers to implement efficient water allocation strategies. 

Computer vision and machine learning have made strides in identifying specific systems like center pivots \cite{geus2020fast,saraiva2020automatic,rodrigues2020circular} but the broader task of irrigation mapping demands nuanced analysis. 
In the western U.S. and other dry regions of the world, a variety of irrigation methods coexist on the landscape. 
In an irrigated watershed of southern Idaho, Bjorneberg et al. found that the
proportion of the agricultural land irrigated with sprinklers increased from ~6\% in~2006
to~59\% in~2016 as furrow irrigated fields were converted to sprinkler systems \cite{bjorneberg2020watershed}. 
Evaluating the effect of different irrigation methods on basin-scale processes requires 
accurate information on the location and distribution of the types of irrigation used 
on‐farm across the irrigated landscape. Currently available irrigation mapping products
\cite{pervez2010mapping,salmon2015global,siebert2005development,siebert2015global}
distinguish between irrigated and non‐irrigated areas at various spatial scales
without explicit information on the methods of irrigation. Regional-scale irrigation
mapping from remotely sensed data has been the object of many studies \cite{pervez2010mapping,bjorneberg2020watershed,bazzi2019mapping}. 
\vspace{-0.05cm}
\section{Description of Data Source}

\textbf{General Information.}
Our multi-resolution dataset IrrMap consists of around one million patches containing LandSat and Sentinel imagery. The dataset includes pixel-precise irrigation mapping, land use, and crop cover annotations at both 30m 
and 10m resolutions while preserving the geolocation of each patch (Table \ref{tab:summary_data}). 
Our dataset covers four states—Arizona, Colorado, Utah, 
and Washington—spanning approximately 1,443,492.31 acre of U.S. agricultural lands (shown in Figure \ref{fig:coverage_area} and Table \ref{tab:dataset_overview}). 

\noindent
\textbf{Remote Sensing Data Source.}
LandSat-8 and Sentinel-2A imagery were obtained from \href{https://earthexplorer.usgs.gov/}{USGS Earth Explorer} for Utah, Arizona, Washington, and Colorado, covering their respective study periods. LandSat-8 provides 30m multispectral and 100m thermal resolution with a 16-day revisit cycle, while Sentinel-2A offers 10m visible and near-infrared resolution with a 5-day revisit cycle. Data acquisition focused on the irrigation season (March–September), a key period for assessing water use and crop conditions, as imagery from other periods may capture snow cover, bare soil, or dormant vegetation, making it less useful for irrigation analysis. Images exceeding 5\% cloud cover, snow, or poor quality—identified via the Quality Assessment (QA) band—were excluded.

\noindent
\textbf{Irrigation Data Source.} Since \emph{irrigation label data} is not readily available from a single source for all states, we collected data for four states from different sources. For Utah (2021–2023), we obtained data from the {Water-Related Land Use (WRLU)}\footnote{\url{https://dwre-utahdnr.opendata.arcgis.com/pages/wrlu-data}} dataset. Washington's irrigation data (2015–2020) was sourced from the Washington State Department of Agriculture Agricultural Land Use dataset (WSDA)\footnote{\url{https://agr.wa.gov/departments/land-and-water/natural-resources/agricultural-land-use}}, while Colorado's data (2018–2020) was acquired from the Colorado Decision Support System (CDSS)\footnote{\url{https://dwr.colorado.gov/services/data-information/gis}}, which provides Geographic Information System (GIS) datasets for all river basins. For Arizona (2015–2017), we utilized the USGS Verified Irrigated Agricultural Lands dataset\footnote{\url{https://catalog.data.gov/dataset/verified-irrigated-agricultural-lands-for-the-united-states-200217}}, a GIS geodatabase developed collaboratively by USGS and the University of Wisconsin. All of these datasets include various land use details, such as vector polygons of irrigated fields, irrigation methods, crop types, water sources, and acreage information across different years. To standardize the analysis across data sources, we mapped various irrigation practices to three primary methods: drip, sprinkler, and flood irrigation (see Table \ref{tab:irr_mapping}). For example, ``Center Pivot'', ``Wheeler Sprinkler'', and ``Traveling Gun'' all fall under sprinkler irrigation. In addition, some labels are noisy as multiple irrigation methods are assigned to a single entry (e.g., ``Drip/Rill/Sprinkler'' or ``Center Pivot/Drip/Sprinkler''). To maintain clarity, such noisy labels are removed, ensuring clean, structured data for reliable analysis. Irrigation methods vary by state; Utah's 2.7 million acres have 0.08\% drip, 34\% sprinkler, and 31\% flood, while Arizona's 566,340 acres include 8\% drip, 46\% flood, and 20\% sprinkler irrigation (See Table \ref{tab:summary_data}).


\noindent
\textbf{Crop Data Source.} For {Crop Data}, A total of 143 distinct crop types were identified across the four states from USGS Verified Irrigated Agricultural Lands datasets. To standardize the analysis, we consolidated these crops into 20 categorical groups based on classifications from Leff et al. \cite{leff2004geographic} and definition from the IR4-Project of U.S. Department of Agriculture (USDA) \cite{ir2012index}. The groups are: Alfalfa, Cereals, Cover Crop, Fibres, Fruits, Grass, Green House, Herb Group, Horticulture, Nursery, Nuts, Oil-bearing crops, Orchard, Pulses, Roots and Tubers, Shrub Plants, Sugar Crops, Vegetables, Vineyard, and an additional category for unspecified crops. For instance, the cereal group includes barley, corn, wheat, and sorghum, while the fruit group encompasses apples, berries, citrus, and melons. The complete mapping of individual crops to their respective groups is provided in the Appendix.

\noindent
\textbf{Land-Use Data Source.} For \emph{Agricultural Land Use Data}, the study area encompasses various land use categories including irrigation, dry agriculture, idle, riparian, sub-irrigation, urban, urban grass, water, and wet flats. From these categories, we focused specifically on irrigated land and agricultural land, as these represent the primary zones requiring active irrigation management.




\begin{table}[h]
    \centering
    \setlength{\tabcolsep}{3pt} 
    \renewcommand{\arraystretch}{1.1} 
    \caption{IrriMap Dataset Distribution Across Western U.S. States for Sentinel and LandSat Imagery. The dataset is divided into training and testing splits for machine learning applications.}
    \label{tab:dataset_distribution}
    \begin{tabular}{|l|c|cccc|c|}
        \hline
        \multirow{2}{*}{\textbf{Source}} & \multirow{2}{*}{\textbf{Split}} & \multicolumn{4}{c|}{\textbf{Regions}} & \multirow{2}{*}{\textbf{IrrMap}} \\
        \cline{3-6}
        & & \textbf{AZ} & \textbf{UT} & \textbf{WA} & \textbf{CO} &  \\
        \hline
        \multirow{3}{*}{\textbf{LandSat}}  
        & Train & 2,655 & 40,224 & 26,852 & 18,284 & 88,015 \\
        & Test  & 665   & 10,057 & 6,716  & 4,573  & 22,011 \\
        \cline{2-7}
        & \textbf{Total} & \textbf{3,320} & \textbf{50,281} & \textbf{33,568} & \textbf{22,857} & \textbf{110,026}  \\
        \hline
        \multirow{3}{*}{\textbf{Sentinel}}  
        & Train & 7,930 & 306,767 & 262,206 & 269,596 & 846,499 \\
        & Test  & 1,985  & 76,693  & 65,555  & 67,401  & 211,634 \\
        \cline{2-7}
        & \textbf{Total} & \textbf{9,915}  & \textbf{383,460}  & \textbf{327,761}  & \textbf{336,997} & \textbf{1,058,133}  \\
        \hline
    \end{tabular}
\end{table}
\begin{table}[h]
\small
    \centering
    \renewcommand{\arraystretch}{1.3}
     \caption{Irrigation Method Mapping}
    \begin{tabular}{|p{6cm}|l|}
        \hline
        \textbf{Original Label} & \textbf{Mapped Label} \\
        \hline
        Drip Microirrigation, Micro-Drip, DRIP, Drip & Drip \\
        \hline
        Traveler Sprinkler, Center Pivot - Tow, Solid State Sprinkler, Overhead, Traveling Gun, Pivot, 
        Traveling Gun, pivot, sprinkler, Center Pivot, Micro-Sprinkler, Micro-Bubbler, Mirco-Bubbler, 
        Sprinkler \& Bubbler, Lateral, Side Roll, Lateral Sprinkler, Other Sprinkler, Microsprinkler, 
        Big Gun, Wheel Line, Big Gun/Sprinkler, Sprinkler/Wheel Line, Big Gun/Center Pivot, 
        Center Pivot/Sprinkler, Center Pivot/Wheel Line, Center Pivot/Sprinkler/Wheel Line, 
        Big Gun/Wheel Line, Wheel line, SPRINKLER, Sprinkler & \multirow{8}{*}{Sprinkler} \\
        \hline
        Furrow, Grated\_Pipe, Improved Flood, Rill, Hand/Rill, None/Rill, Grated\_pipe, Gated\_pipe, 
        FLOOD, FURROW, GATED\_PIPE & \multirow{3}{*}{Flood} \\
        \hline
        Not Specified, Micro, No Label, Research, Uncertain, 
        Drip/None, Hand, Big Gun/Drip, Drip/Big Gun, Drip/Rill/Sprinkler, Rill/Sprinkler, 
        Drip/Micro-Sprinkler, Drip/Wheel Line, Center Pivot/Rill, Rill/Wheel Line, Drip/Rill, 
        Center Pivot/None, Center Pivot/Rill/Wheel Line, Center Pivot/Rill/Sprinkler, Rill/Sprinkler/Wheel Line, 
        Center Pivot/Drip, Hand/Sprinkler, Drip/Sprinkler, Sub-irrigated, Dry Crop, Sprinkler And Drip, 
        Center Pivot/Drip/Sprinkler, None/Sprinkler/Wheel Line,  Unknown & \multirow{8}{*}{Removed} \\
        \hline
    \end{tabular}
    \vspace{-0.3cm}
    \label{tab:irr_mapping}
\end{table}

\section{ML-ready Dataset Preparation}

We create a structured (consistent formats, encoded categorical values, and 
normalized numerical data) and preprocessed (spatially consistent, 
quality-checked, feature engineered, and carefully partitioned into train/test) 
ML-ready dataset (IrrMap) for seamless application of ML models as follows: 

\noindent
\textbf{Complete Data Processing Pipeline.} 
To streamline the dataset preparation process, we developed a pipeline that performs preprocessing, patch segmentation, label integration, and train-test splitting as shown in Figure \ref{fig:irrMap}. Through this pipeline, we have streamlined \textbf{25.61 TB} of raw satellite data into a well-structured \textbf{6.2 TB} IrrMap dataset, significantly enhancing accessibility and usability of the dataset. The entire pipeline is built using GDAL, Rasterio, OpenCV, and PyTorch Lightning, ensuring efficiency and reproducibility. In addition, we provide a PyTorch Lightning DataModule that allows seamless integration with deep learning models.
Now, we describe each step of the pipeline.

\noindent
\textbf{Data acquisition.}
We acquired 2,407 and 36,638 image tiles from LandSat and Sentinel respectively, totaling~25.61 TB~(2.78 TB for LandSat and 22.83 TB for Sentinel) of data and covering 
over 11 million acres of irrigated land in the western United States.


\noindent
\textbf{Geospatial Standardization.} We first reproject each LandSat and 
Sentinel image to the WGS-84/EPSG:4326 coordinate system to ensure 
uniformity across the collected satellite images from different 
regions~\cite{slater1998wgs}. This facilitates accurate overlaying with 
ground truth label from different sources (e.g., crop labels, irrigation 
labels, land use labels, etc.).

\noindent
\textbf{Patch Segmentation.} Following standard practices, we divide each image into $224\times 224$ non-overlapping patches to reduce computational complexity while maintaining spatial integrity. Each patch covers approximately 45 square kilometers at 30m resolution for LandSat and 5 square kilometers at 10m resolution for Sentinel, ensuring a balance between capturing relevant spatial information and offering efficient deep learning model training. Moreover, we discard all patches containing poor-quality pixels to ensure data quality and reliability (based on QA-Bands).

\noindent
\textbf{RGB Images Creation.} We extract the Red, Green, and Blue bands
from satellite imagery (e.g., LandSat or Sentinel) and stack ($224\times 
224\times 3$) them to form a true-color composite. In addition, we apply 
the standard step of gamma correction (a nonlinear adjustment) that modifies
image brightness and contrast by altering pixel intensity values for improved visibility \cite{rahman2016adaptive}).

\noindent
\textbf{Derived Indices Creation.} To enhance surface property analysis for irrigation mapping, we compute various spectral indices capturing vegetation health\footnote{https://www.nv5geospatialsoftware.com/docs/AlphabeticalListSpectralIndices.html}, soil conditions, and water availability. NDVI, GNDVI, and CIgreen assess vegetation vigor and chlorophyll content, while EVI, SAVI, and MSAVI adjust for atmospheric and soil background effects, improving vegetation monitoring in different environments \cite{meivel2022monitoring,tucker1985satellite,huete2002overview}. NDWI detects water bodies and moisture content, aiding in drought assessment, whereas NDTI distinguishes tilled from untilled soils for land management analysis \cite{gao1996ndwi,zheng2014remote}. PRI measures plant stress and photosynthetic efficiency, OSAVI refines soil adjustment for vegetation detection, and WDRVI enhances differentiation in high biomass areas \cite{gitelson2004wide,fern2018suitability,thenot2002photochemical}. RVI provides an alternative vegetation measure with reduced sensitivity to atmospheric variations \cite{gonenc2019comparison}. Together, these indices offer valuable insights for irrigation mapping, agricultural monitoring, and environmental assessment (More details in Appendix.

\noindent
\textbf{Label Integration.} We begin by projecting all label data, including crop, irrigation, and land use information, into the WGS-84 coordinate system (EPSG:4326) to ensure accurate spatial alignment with our satellite image patches. A pixel-wise labeling approach is then employed to integrate these labels into a unified structure. This structured integration enables a precise and consistent representation of irrigation, crop type, and land use information while ensuring data reliability and spatial coherence.
\begin{itemize}
    \item \textbf{Irrigation Mask.} Each patch is assigned an irrigation mask $ Y \in \{0,1,2,3\}^{224 \times 224} $, where each pixel is mapped to one of four irrigation methods: 0 (Others/No-Irrigation), 1 (Flood), 2 (Sprinkler), and 3 (Drip).  

    \item \textbf{Crop Mask.} We generate a crop mask $ C \in \{0,1,\dots,21\}^{224 \times 224} $, where each pixel is classified into one of 21 crop types, with 0 indicating no crop presence.  

    \item \textbf{Land Mask.} A binary land mask, $ L \in \{0,1\}^{224 \times 224} $, is derived from land use data, where 1 represents agricultural land and 0 denotes non-agricultural areas, ensuring spatial integrity.

\end{itemize}

\begin{figure}
    \centering
    \includegraphics[width=.9\linewidth]{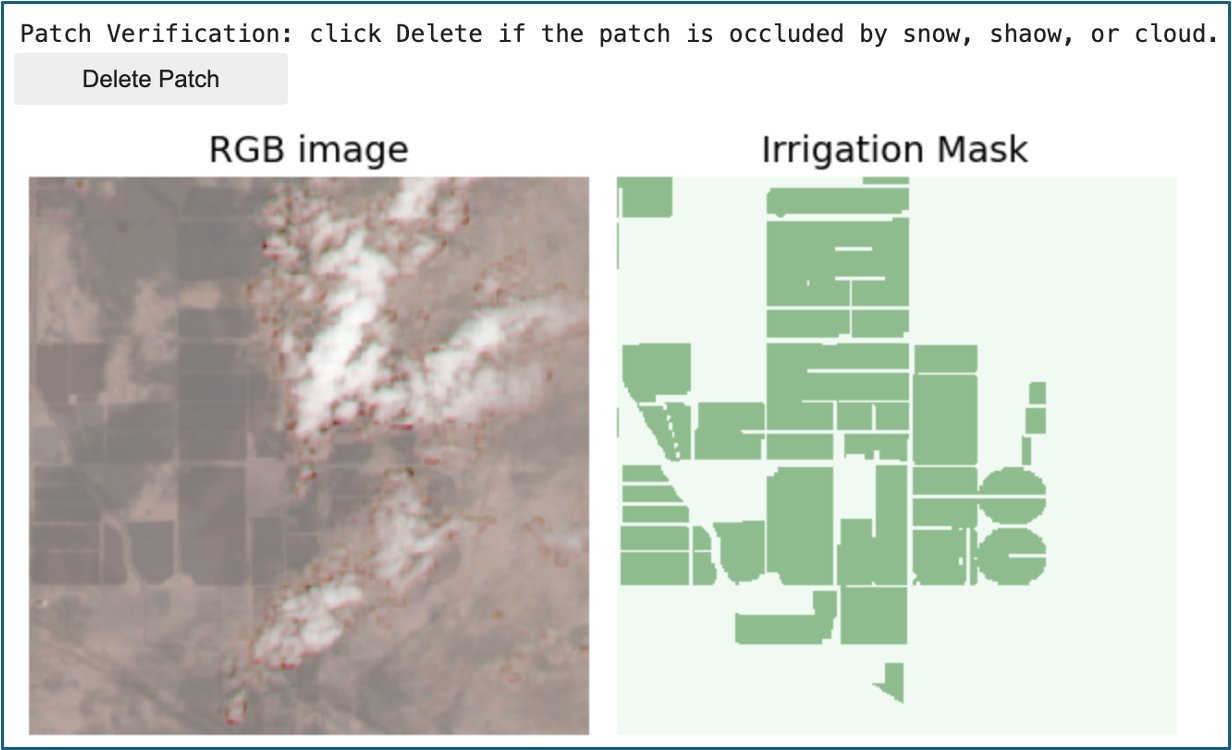}
    \caption{Our designed user interface for quality filtering at the patch level.}
    \label{fig:ui-filter}
\end{figure}

\noindent
\textbf{Quality Filtering.} Since the majority of created patches lack valid irrigation data, 
we exclude patches where more than~99.99\% of pixels are labeled as non-irrigated. This 
ensures that each selected patch contains meaningful irrigation information for analysis. 
After quality filtering, we find a total of 110,026 patches for LandSat imagery 
and~1,058,133 patches for Sentinel~(shown in Table~\ref{tab:dataset_distribution}). Even 
after following standard tile-level filtering processes to remove imagery affected by snow, clouds, 
or shadows~(as guided by the data quality band), it is 
insufficient to capture subtle occlusions at the patch-level such as thin cloud veils, shadow fringes, and 
edge artifacts near field boundaries~(see supplement for examples). For instance, when clouds or shadows affect only a subset of pixels within a tile, the tile-level quality assessment may not trigger exclusion criteria, yet patches sampled from the degraded regions will contain corrupted pixel values that compromise downstream processing algorithms. 
These quality issues, 
although infrequent, significantly degrade classification accuracy due to their 
confounding spectral signals. Therefore, we manually reviewed and annotated over~18,000 
patches to ensure high-fidelity supervision. Each patch in our dataset is assigned a 
binary data quality label$~Q \in \{0,1\} $, where~$0$ indicates contamination due to 
clouds, shadows, or snow and~$1$ represents a clean image. We designed a user interface to manually filter the patches that are occluded by snow, shadow, or cloud (shown in Figure~\ref{fig:ui-filter}). This dual-layer 
filtering~(automated plus manual) ensures dataset integrity and provides a foundation 
for developing future models to learn from both clean and noisy samples.


\noindent
\textbf{Train-Test Splitting.} To ensure a robust and unbiased
evaluation of machine learning models, we carefully split the IrriMap dataset into
80\% training and 20\% testing sets. As irrigation practices vary
significantly vary across different geographical regions (e.g., Utah 
has only 0.08\% drip-irrigated lands in the labeled data), a state-wise
stratified split is performed. Each state's dataset is independently divided into an 80-20\% split, 
and the resulting 20\% testing portions are then combined to form a spatially well-represented 
test set, thereby mitigating geographic biases. State-wise distribution of training and testing 
datasets is shown in Table \ref{tab:dataset_distribution}. A total of 934,514 (88,015 for LandSat and 846,499 for Sentinel) patches are created for the training dataset. 
The test dataset has 233,645~(22,011 for LandSat and 211,634 for Sentinel) patches in total.
To improve accessibility for downstream users, we additionally provide a lightweight version
of the training set, containing~234,861 Sentinel and~26,316 LandSat patches. These subsets 
were created by randomly sampling~30\% of the unique satellite scene tiles and retaining all 
associated patches.

\begin{figure*}
    \centering
    \includegraphics[width=.85\linewidth]{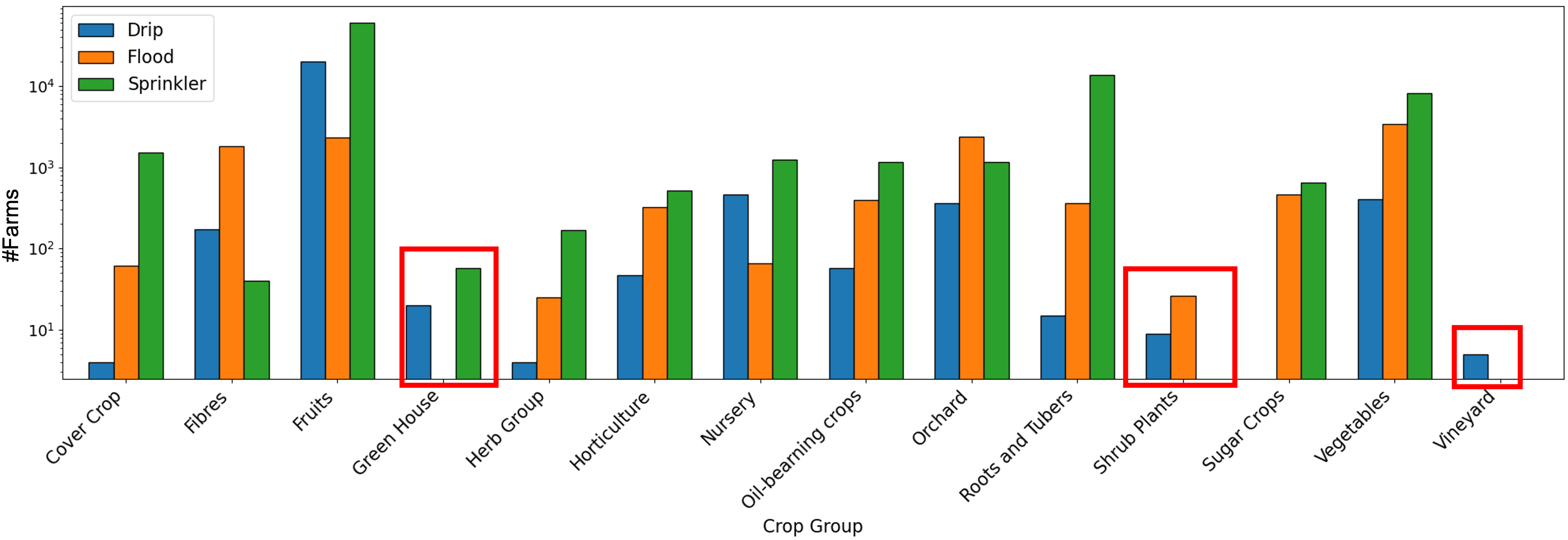}
    \caption{Distribution of irrigation methods (Drip, Flood, and Sprinkler) across different crop groups for the studied regions. Y-axis denotes the number of farms in log-scale. Some crop groups, such as `Green House' and `Vineyard' do not use flood irrigation, whereas `Shrub Plants' cultivated farms do not use sprinkler irrigation. Noteably, `Nursery'  and `Fruit' cultivated farms prefer drip irrigation more than flood irrigation. Sprinkler irrigation is popular in `Root and Tube' cultivated farms across the studied regions.}
    \label{fig:crop-distribution}
\end{figure*}
\section{Data Analysis}
\textbf{Irrigation-Crop Relationship.} Our analysis of irrigation methods across farms highlights distinct preferences based on crop groups in Figure \ref{fig:crop-distribution}. We notice that vineyards rely on drip irrigation only, while nursery and orchard farms generally use drip irrigation. This analysis emphasizes the need for precise water control in high-value crops. Fruit and root crops are mainly irrigated using sprinkler systems, which provide uniform water distribution, whereas fiber crops, such as cotton, depend heavily on flood irrigation. Notably, greenhouse farms avoid flood irrigation and opt for more controlled sprinkler and drip methods, while shrub plant farms do not use sprinkler irrigation. These findings underscore the relationship between the crop types and irrigation methods.

\noindent
\textbf{Irrigation Diversity.} To understand the diversity of irrigation practices within the dataset, we quantify diversity using Shannon Index \cite{marcon2014generalization}. For each patch, we calculate the Shannon diversity index $H$ for each patch as follows:
\begin{equation}
    H = - \sum_i \frac{p_i}{\sum_j p_j} \times \ln \left(\frac{p_i}{\sum_j p_j} \right)
\end{equation}

which $p_i$ represents the fraction of area occupied by the irrigation method $i$ and $i,j \in \{\text{Drip, Sprinkler, Flood}\}$. If a patch contains only one irrigation method, the Shannon index is 0 (indicates homogeneous), whereas a patch with mixed irrigation has a higher diversity score. Note that the score is highest when all three (drip, sprinkler, and flood) have equally irrigated lands within the patch. 

\begin{figure}[th]
    \centering
    \includegraphics[width=\linewidth]{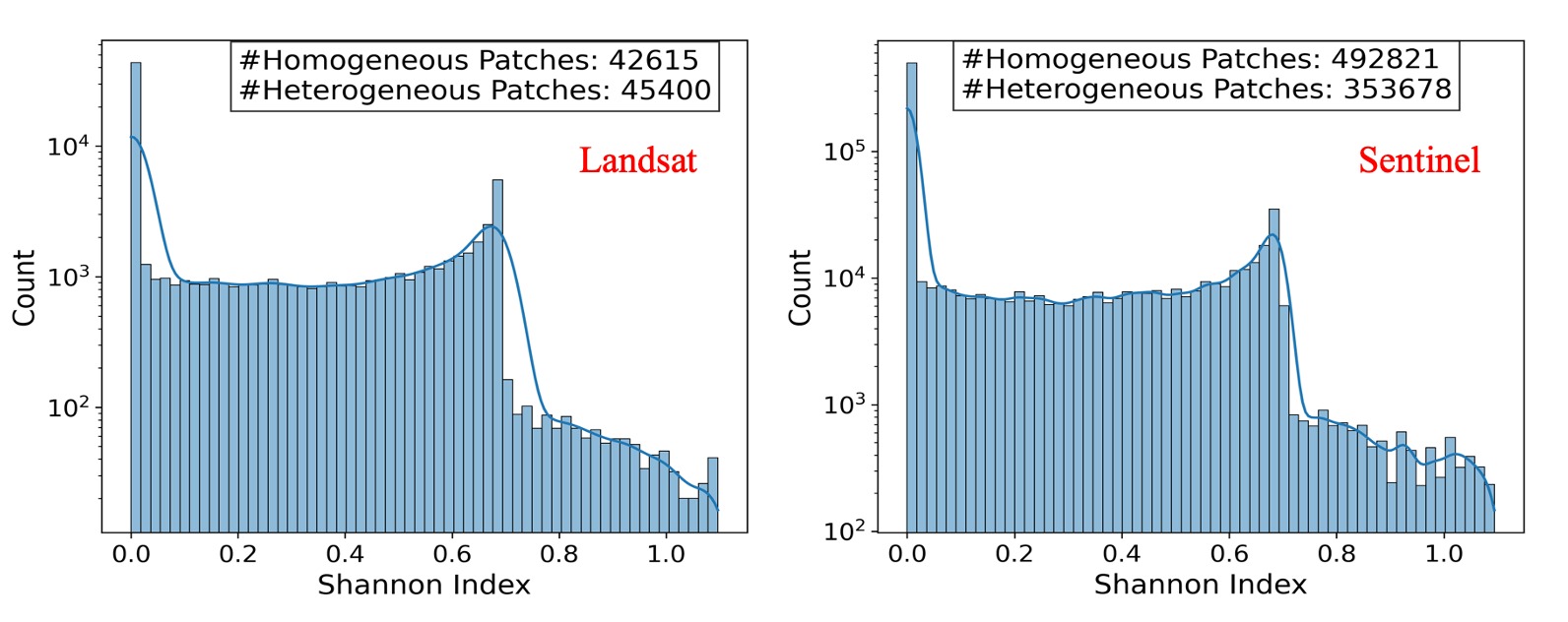}
    \caption{The figure shows irrigation patterns for the IrrMap dataset through Shannon diversity indices. The left figure presents the distribution for the LandSat data, whereas the right figure displays for the Sentinel data. Two distinct spatial patterns are noticed: both single-method dominance in some patches and mixed irrigation practices (index $0.6$) in others.}
    \vspace{-0.1cm}
    \label{fig:shannon-diversity}
\end{figure}

The varying complexity of irrigation mapping in the dataset is evident in the Shannon diversity indices of our dataset patches (as shown in Figure \ref{fig:shannon-diversity}). We notice that the IrrMap dataset (both LandSat and Sentinel) have similar pattern, with a high concentration of low-diversity patches and a secondary peak near 0.7 (indicates diverse irrigation practices). We further show that a significant portion of the dataset consists of homoegeneous patches (42,615 for LandSat and 492,821 for Sentinel) where a single irrigation method is only available within the patch. However, LandSat data has 48\% homogeneous patches compared to 58\% of homogeneous patches for Sentinel, suggesting a higher representation of diversely irrigated lands for LanSat. As Sentinel captures 10-meter-resolution images, the patches zoom into the farmlands and provide finer spatial details. This leads to a higher proportion of homogeneous patches. In contrast, LandSat patches cover a broder area (45 $km^2$ per patch), capturing multiple irrigation methods within a patch. It is worth noting that patches from Sentinel provide finer details with low coverage, leading to increased training and evaluation costs, whereas LandSat patches cover more area with higher diversity, leading to challenges for correctly identifying irrigation methods from a machine learning perspective.

\begin{figure*}
    
\end{figure*}

\section{Benchmarking}
\begin{table*}[!ht]
    \centering
    \footnotesize
    \renewcommand{\arraystretch}{1.2}
    \caption{Irrigation Classification Performance for Different Input Layers, reporting Precision (P), Recall (R), F1-score (F), and Intersection over Union (IoU) for Flood, Sprinkler, and Drip irrigation types across different states.}
    \begin{tabular}{|l|l|cccc|cccc|cccc|}
        \hline
        \textbf{State} & \textbf{Input Layers} & \multicolumn{4}{c|}{\textbf{Flood}} & \multicolumn{4}{c|}{\textbf{Sprinkler}} & \multicolumn{4}{c|}{\textbf{Drip}} \\
        \cline{3-14}
        & & \textbf{P} & \textbf{R} & \textbf{F} & \textbf{IoU} 
        & \textbf{P} & \textbf{R} & \textbf{F} & \textbf{IoU} 
        & \textbf{P} & \textbf{R} & \textbf{F} & \textbf{IoU} \\
        \hline
        \multirow{4}{*}{AZ} 
        & RGB        & 0.5937 & 0.4673 & 0.5230 & 0.3541 & 0.5744 & 0.4247 & 0.4883 & 0.3231 & 0.4744 & 0.3604 & 0.4096 & 0.2576 \\
        & RGB+CROP   & \textbf{0.7359} & \textbf{0.8093} & \textbf{0.7708} & \textbf{0.6271} & \textbf{0.7070} & \textbf{0.7473} & \textbf{0.7266} & \textbf{0.5706} & \textbf{0.6821} & \textbf{0.6306} & \textbf{0.6554} & \textbf{0.4874} \\
        & RGB+LAND   & 0.6589 & 0.7588 & 0.7054 & 0.5448 & 0.6700 & 0.6628 & 0.6664 & 0.4997 & 0.6744 & 0.4089 & 0.5091 & 0.3415 \\
        & RGB+NDVI   & 0.5966 & 0.5653 & 0.5805 & 0.4090 & 0.6288 & 0.4016 & 0.4901 & 0.3246 & 0.4745 & 0.3977 & 0.4327 & 0.2761 \\
        \hline
        \multirow{4}{*}{CO} 
        & RGB        & 0.6108 & 0.2555 & 0.3603 & 0.2197 & 0.5857 & 0.4111 & 0.4831 & 0.3185 & 0.0000 & 0.0000 & 0.0000 & 0.0000 \\
        & RGB+CROP   & \textbf{0.6909} & \textbf{0.6477} & \textbf{0.6686} & \textbf{0.5022} & \textbf{0.7251} & \textbf{0.6695} & \textbf{0.6962} & \textbf{0.5340} & \textbf{0.2798} & \textbf{0.1653} & \textbf{0.2078} & \textbf{0.1159} \\
        & RGB+LAND   & 0.6648 & 0.5497 & 0.6018 & 0.4304 & 0.6701 & 0.5926 & 0.6290 & 0.4588 & 0.2659 & 0.0239 & 0.0438 & 0.0224 \\
        & RGB+NDVI   & 0.5660 & 0.3979 & 0.4673 & 0.3049 & 0.6415 & 0.3752 & 0.4734 & 0.3101 & 0.0000 & 0.0000 & 0.0000 & 0.0000 \\
        \hline
        \multirow{4}{*}{UT} 
        & RGB        & 0.6215 & 0.3030 & 0.4074 & 0.2558 & 0.5307 & 0.5341 & 0.5324 & 0.3628 & 0.0000 & 0.0000 & 0.0000 & 0.0000 \\
        & RGB+CROP   & \textbf{0.6741} & \textbf{0.6549} & \textbf{0.6644} & \textbf{0.4974} & \textbf{0.6870} & \textbf{0.7138} & \textbf{0.7001} & \textbf{0.5386} & \textbf{0.4221} & \textbf{0.0139} & \textbf{0.0269} & \textbf{0.0136} \\
        & RGB+LAND   & 0.6878 & 0.6427 & 0.6645 & 0.4976 & 0.6746 & 0.6579 & 0.6661 & 0.4994 & 0.1530 & 0.0028 & 0.0055 & 0.0028 \\
        & RGB+NDVI   & 0.5909 & 0.4135 & 0.4866 & 0.3215 & 0.5927 & 0.4598 & 0.5179 & 0.3494 & 0.0000 & 0.0000 & 0.0000 & 0.0000 \\
        \hline
        \multirow{4}{*}{WA} 
        & RGB        & 0.4503 & 0.1304 & 0.2023 & 0.1125 & 0.5693 & 0.4607 & 0.5093 & 0.3416 & 0.3766 & 0.0949 & 0.1516 & 0.0820 \\
        & RGB+CROP   & \textbf{0.5926} & \textbf{0.4548} & \textbf{0.5146} & \textbf{0.3464} & \textbf{0.7237} & \textbf{0.7725} & \textbf{0.7473} & \textbf{0.5966} & \textbf{0.4899} & \textbf{0.5068} & \textbf{0.4982} & \textbf{0.3317} \\
        & RGB+LAND   & 0.5256 & 0.4264 & 0.4709 & 0.3079 & 0.7013 & 0.7316 & 0.7161 & 0.5577 & 0.5283 & 0.3326 & 0.4082 & 0.2564 \\
        & RGB+NDVI   & 0.3794 & 0.2458 & 0.2983 & 0.1753 & 0.5872 & 0.4490 & 0.5089 & 0.3413 & 0.4147 & 0.1238 & 0.1907 & 0.1054 \\
        \hline
    \end{tabular}
    \label{tab:irrigation_performance}
\end{table*}

\begin{table*}[!ht]
\footnotesize
    \centering
    \renewcommand{\arraystretch}{1.2}
    \caption{Irrigation Classification Performance for Different Input layers, reporting Precision (P), Recall (R), F1-score (F), and Intersection over Union (IoU) for Flood, Sprinkler, and Drip irrigation methods across LandSat and Sentinel data.}
    \begin{tabular}{|l|l|cccc|cccc|cccc|}
        \hline
        \textbf{Satellite} & \textbf{Input Layers} & \multicolumn{4}{c|}{\textbf{Flood}} & \multicolumn{4}{c|}{\textbf{Sprinkler}} & \multicolumn{4}{c|}{\textbf{Drip}} \\
        \cline{3-14}
        & & \textbf{P} & \textbf{R} & \textbf{F} & \textbf{IoU} 
        & \textbf{P} & \textbf{R} & \textbf{F} & \textbf{IoU} 
        & \textbf{P} & \textbf{R} & \textbf{F} & \textbf{IoU} \\
        \hline
        \multirow{4}{*}{LandSat} 
        & RGB        & 0.6080 & 0.2794 & 0.3829 & 0.2368 & 0.5666 & 0.4540 & 0.5041 & 0.3369 & 0.4133 & 0.3322 & 0.4145 & 0.2614 \\
        & RGB+CROP   & \textbf{0.6821} & \textbf{0.6484} & \textbf{0.6648} & \textbf{0.4979} & \textbf{0.7176} & \textbf{0.7236} & \textbf{0.7206} & \textbf{0.5633} & 0.5166 & \textbf{0.5100} & \textbf{0.5133} & \textbf{0.3452} \\
        & RGB+LAND   & 0.6671 & 0.5922 & 0.6274 & 0.4571 & 0.6857 & 0.6661 & 0.6758 & 0.5103 & \textbf{0.5509} & 0.3322 & 0.4145 & 0.2614 \\
        & RGB+NDVI   & 0.5685 & 0.4047 & 0.4728 & 0.3096 & 0.6058 & 0.4224 & 0.4977 & 0.3313 & 0.4360 & 0.1629 & 0.2372 & 0.1346 \\
        \hline
        \multirow{4}{*}{Sentinel} 
        & RGB        & 0.5497 & 0.4649 & 0.5038 & 0.3367 & 0.7488 & 0.6028 & 0.6679 & 0.5014 & 0.5775 & 0.0283 & 0.0539 & 0.0277 \\
        & RGB+CROP   & \textbf{0.7259} & \textbf{0.6942} & \textbf{0.7097} & \textbf{0.5500} & \textbf{0.8061} & \textbf{0.8290} & \textbf{0.8174} & \textbf{0.6911} & 0.5675 & \textbf{0.6558} & \textbf{0.6085} & \textbf{0.4373} \\
        & RGB+LAND   & 0.6861 & 0.6449 & 0.6649 & 0.4980 & 0.7947 & 0.7547 & 0.7742 & 0.6316 & \textbf{0.5926} & 0.3936 & 0.4731 & 0.3098 \\
        & RGB+NDVI   & 0.6172 & 0.4137 & 0.4954 & 0.3292 & 0.7252 & 0.6879 & 0.7061 & 0.5457 & 0.4827 & 0.2039 & 0.2867 & 0.1673 \\
        \hline
    \end{tabular}
    \label{tab:irrigation_performance_both}
\end{table*}

\textbf{Evaluation Setup. }To evaluate the effectiveness of IrrMap for irrigation method classification, we design an experiment using satellite imagery from LandSat. 
In particular, we demonstrate how additional masks can contribute to performance improvements
as well as quantify the impact of each of them. We select three primary layers (CROP, LAND and NDVI) and explore the extent to which each one enhanced the results. 

We classify irrigation types—Flood, Sprinkler, and Drip—by leveraging different input layers and benchmarking their performance across various configurations. We employed a UNet-style ResNet-34 architecture for all baseline experiments, training the models on the complete LandSat training dataset and presenting results on the test set of each state. We train the model for 10 epochs on an NVIDIA A100 80GB GPU. The motivation behind this experiment is to assess how incorporating auxiliary layers—such as NDVI, Crop Mask, and Land Mask—alongside RGB imagery improves impact the performance. More hyper-parameter settings and results on the Sentinel data are present in Appendix.
\\
\\
\noindent
\textbf{Results. }The results in Table~\ref{tab:irrigation_performance} demonstrate that incorporating additional feature layers alongside RGB imagery leads to varying degrees of improvement in irrigation classification performance. RGB + Crop Mask consistently achieves the highest accuracy, with an average F1-score improvement of 30–50\% over RGB alone, particularly excelling in Flood and Sprinkler irrigation. The Land Mask augmentation also enhances classification, yielding a 20–40\% increase in IoU and F1-score compared to RGB-only, especially for Flood irrigation. Meanwhile, NDVI augmentation provides moderate benefits, with a 10–25\% boost in F1-score and IoU for Sprinkler and Flood irrigation methods but does not perform as well for Drip irrigation. RGB-only models still produce reasonable results, particularly for Flood irrigation, but exhibit relatively lower performance in distinguishing Sprinkler and Drip irrigation methods. These findings suggest that crop-specific information (via Crop Mask) is the most informative additional layer, while land use and vegetation indices (NDVI) contribute moderately, depending on the irrigation methods. This opens up an interesting research direction on how to best leverage these additional layers—whether through feature fusion techniques, advanced model architectures, or task-specific weighting—to achieve optimal performance.

In Table \ref{tab:irrigation_performance_both}, we present the overall performance on the training set. For training on Sentinel data, we use the same hyper-parameters and model architecture as LandSat. We randomly sample 100K training and test data from the IrrMap Sentinel dataset and train the model for five epochs.

The results indicate that the inclusion of additional bands, such as crop type and NDVI, has a notable impact on irrigation classification performance for both LandSat and Sentinel datasets. In LandSat, the RGB+CROP layers consistently outperforms other configurations, particularly in Flood and Sprinkler irrigation, suggesting that crop type information significantly enhances classification accuracy. Similarly, RGB+NDVI improves performance compared to standard RGB, but it does not surpass RGB+CROP, highlighting the greater influence of crop data over vegetation indices in LandSat imagery. For Sentinel, the impact of additional bands is even more pronounced, with RGB+CROP achieving the highest F1-scores and IoU across all irrigation methods, particularly for Sprinkler and Drip irrigation, where detailed crop information enhances model differentiation. RGB+LAND has also shown similar impact. The RGB+NDVI configuration also improves performance, especially in Sprinkler irrigation, reinforcing the value of vegetation indices at Sentinel’s higher resolution. Overall, the inclusion of auxiliary bands positively influences classification performance, with crop type data playing a critical role in both datasets.

\section{Potential Applications of the Dataset}
\label{sec:app}
In addition to irrigation mapping, our IrrMap dataset has a wide range of agricultural applications, as follows: 

\begin{itemize}
    \item \textbf{Homogenous and Heterogenous Irrigated Field Identification}: Our large-scale dataset enables classifying irrigation patterns into homogeneous (single practice) and heterogeneous (mixed practice) regions, improving water resource analysis. Additionally, 20K labeled samples of cloud, snow, and shadow enhance quality control models for filtering contaminated satellite imagery, ensuring reliable agricultural monitoring.

    \item \textbf{Temporal Pattern Recognition and Change Detection in Irrigation Over Years}: A trained model using IrrMap, can be used for hindcasting, analyzing past irrigation practices, crop distribution, and land-use changes. By leveraging historical satellite imagery and spectral indices, the model can detect shifts in irrigation methods, variations in soil moisture, and crop rotation patterns. This enables researchers to assess long-term trends in water usage and agricultural adaptation. 

    \item \textbf{Foundation Model for Agricultural Assets}: The large-scale nature of our dataset provides an opportunity to build foundation models for agricultural applications \cite{cong2022satmae,smith2023earthpt,jakubik2310foundation}. The dataset can be used to develop agricultural-specific foundation models through self-supervised pretraining, enabling transfer learning for regions with limited labeled data. Clustering algorithms can identify similar agricultural practices across different regions, while anomaly detection systems can spot unusual irrigation practices.
    
    \item \textbf{Irrigation Mapping through Knowledge-Guided Deep Learning}: Beyond pure machine learning approaches, our dataset supports the development of hybrid systems that combine agricultural domain knowledge with data-driven insights \cite{liu2024knowledge,liu2022kgml}. These models can integrate expert knowledge about crop information, local agricultural practices, and regional climate patterns with machine learning predictions to provide more reliable and practical solutions. Applications include smart irrigation scheduling systems that consider multiple factors (crop type, growth stage, irrigation practice, weather patterns), resource optimization tools for water, and decision support systems for crop rotation planning.
\end{itemize}

In addition, our proposed pipeline (Figure \ref{fig:irrMap}) streamlines the collection, processing, and integration of multi-scale satellite data, significantly reducing the effort required for large-scale geospatial analysis. By leveraging labeled shapefiles, it queries and retrieves satellite imagery (Sentinel and LandSat) from cloud platforms such as Earth Explorer (EE). The pipeline also automates spatial patch generation, seamlessly aligning satellite imagery with auxiliary datasets like climate analysis models and soil structure maps.


\section{Limitation of the dataset}
The IrriMap dataset and trained model are currently limited to the western United States, which may restrict its applicability to regions with different climatic and agricultural conditions. All collected irrigation mapping data sources assume that irrigation practices and crop cultivation remain consistent throughout a season, which may not account for intra-seasonal variations due to environmental or management changes. 
Since the dataset is merged from multiple sources, it is subject to data inconsistencies, 
noise, and rasterization artifacts that may impact accuracy. 
The crop mask relies on the cropland data, which is a model-generated mask. Any errors in 
model predictions can percolate into the classification tasks. Therefore, it is important to not 
over-rely on this dataset as well as substitute it with better equivalent datasets where available. 
Groundwater data is not available in our IrrMap, limiting the model's ability to assess subsurface water interactions. 
While our mapping of original irrigation labels to three broad categories (drip, sprinkler, and flood) facilitates standardized analysis across diverse data sources, we acknowledge that this abstraction may obscure nuanced differences between specific technologies (e.g., center pivot vs. traveling gun). This trade-off was necessary to enable scalable, multi-state analysis, but future iterations of IrrMap will aim to preserve finer-grained irrigation method distinctions to support precision water management research.


\section{Conclusion}
We present IrrMap, a large-scale ML-ready dataset (1.1 million patches) for irrigation mapping by combining multi-resolution satellite imagery with crop, land use, and vegetation indices that covers 14.1 million acres of farmland. The dataset provides a comprehensive view of irrigation method distribution, spatial diversity, and crop-irrigation relationship. Our benchmark results demonstrate that incorporating auxiliary data such as crop, land use, and vegetable indices significantly improves irrigation classification accuracy. Moreover, our pipeline for ML-ready datasets is designed for data integration that enables seamless incorporation of additional data layers to expand analytical capabilities. 

In future work, we plan to expand IrrMap to diverse agricultural regions, particularly semi-arid and developing agricultural landscapes. We aim to integrate temporal analysis using historical and near-real-time satellite imagery to track shifts in irrigation adoption and drought monitoring. In addition, incorporating groundwater data, soil moisture estimates, and real-time climate variables would enable holistic and precise irrigation monitoring, supporting sustainable water resource management.


\begin{acks}
This work was supported in part by the NSF and USDA-NIFA under the AI Institute: Agricultural AI for Transforming Workforce and Decision Support (AgAID) award No.~2021-67021-35344, and by University of Virginia Strategic Investment Fund award number SIF160.
\end{acks}
\bibliographystyle{ACM-Reference-Format}
\balance
\bibliography{sample-base}

\appendix

\onecolumn

\begin{center}
\Large\textbf{Supplementary Material}
\end{center}

More details are available in our online Appendix~\cite{mandal2025irrmap}.
\section{Evaluation Metrics}
We evaluate the irrigation mapping task using four standard segmentation metrics: Intersection over Union (IoU), Precision (P), Recall (R), and F1-Score (F1). Let \( Y \) and \( \overline{Y} \) be the ground truth and predicted masks, respectively, for an image of size \( H \times W \), where each pixel \((i,j)\) is assigned a class \( k \in \mathcal{Y} \). The ground truth and predicted pixel sets for class \( k \) are defined as:
\begin{equation}
    T_k = \{(i, j) \mid Y_{i,j} = k\}, \quad 
    M_k = \{(i, j) \mid \overline{Y}_{i,j} = k\}.
\end{equation}

The evaluation metrics precision, recall, F1, and IoU are computed as:
\begin{align}
    \text{P}_k = \frac{|M_k \cap T_k|}{|M_k|}, \quad \quad \quad
    \text{R}_k &= \frac{|M_k \cap T_k|}{|T_k|}, \quad\quad
    \text{F1}_k = \frac{2 \times P_k \times R_k}{P_k+R_k},
    \quad \quad
    \text{IoU}_k = \frac{|M_k \cap T_k|}{|M_k \cup T_k|},
\end{align}
Precision measures the proportion of correctly predicted irrigated pixels among all predicted as class \( k \), while recall quantifies the fraction of correctly identified irrigated pixels out of all actual class \( k \) pixels. On the contrary, F1-Score computes the harmonic mean of precision and recall. IoU is defined as the ratio of intersection to union, provides a more balanced spatial evaluation by penalizing both over-segmentation (false positives) and under-segmentation (false negatives).

\section{Benchmarking}
\label{sec:result}
\textbf{Parameter Settings. } Hyperparameter details for supervised experiments can be seen in Table~\ref{tab:hyperparams}. We did not perform significant hyperparameter tuning. Model selection was based on the best performance on the validation set.\\
\noindent

\begin{table*}[ht]
\footnotesize
    \centering
    \caption{Hyperparameter Details}
    \label{tab:hyperparams}
    \begin{tabular}{lcccccc}
        \toprule
        \multicolumn{7}{l}{\textbf{Data Setup}} \\
        \midrule
        \textbf{Task Type} & \textbf{Input Dim} & \textbf{RGB} & \textbf{RGB+CROP} & \textbf{RGB+LAND} & \textbf{RGB+NDVI} & \textbf{Output Ch.} \\
        Irrigation Method Segmentation & $224\times224$ & 3 & 24 & 4 & 4 & 4 \\
        \midrule
        \multicolumn{7}{l}{\textbf{Training Setup}} \\
        \midrule
        \textbf{Loss Function} & \textbf{Optimizer} & \textbf{Learning Rate} & \textbf{Epochs (L)} & \textbf{Epochs (S)} & \textbf{Model Sel.} & \textbf{Framework} \\
        Cross Entropy & Adam & $1\times10^{-2}$ & 10 & 5 & Best (val) & PyTorch Lightning \\
        \bottomrule
    \end{tabular}
\end{table*}


\section{Data Analysis}
\begin{figure*}[th]
    \centering
    \begin{subfigure}[b]{0.235\textwidth}
        \centering
        \includegraphics[width=\textwidth]{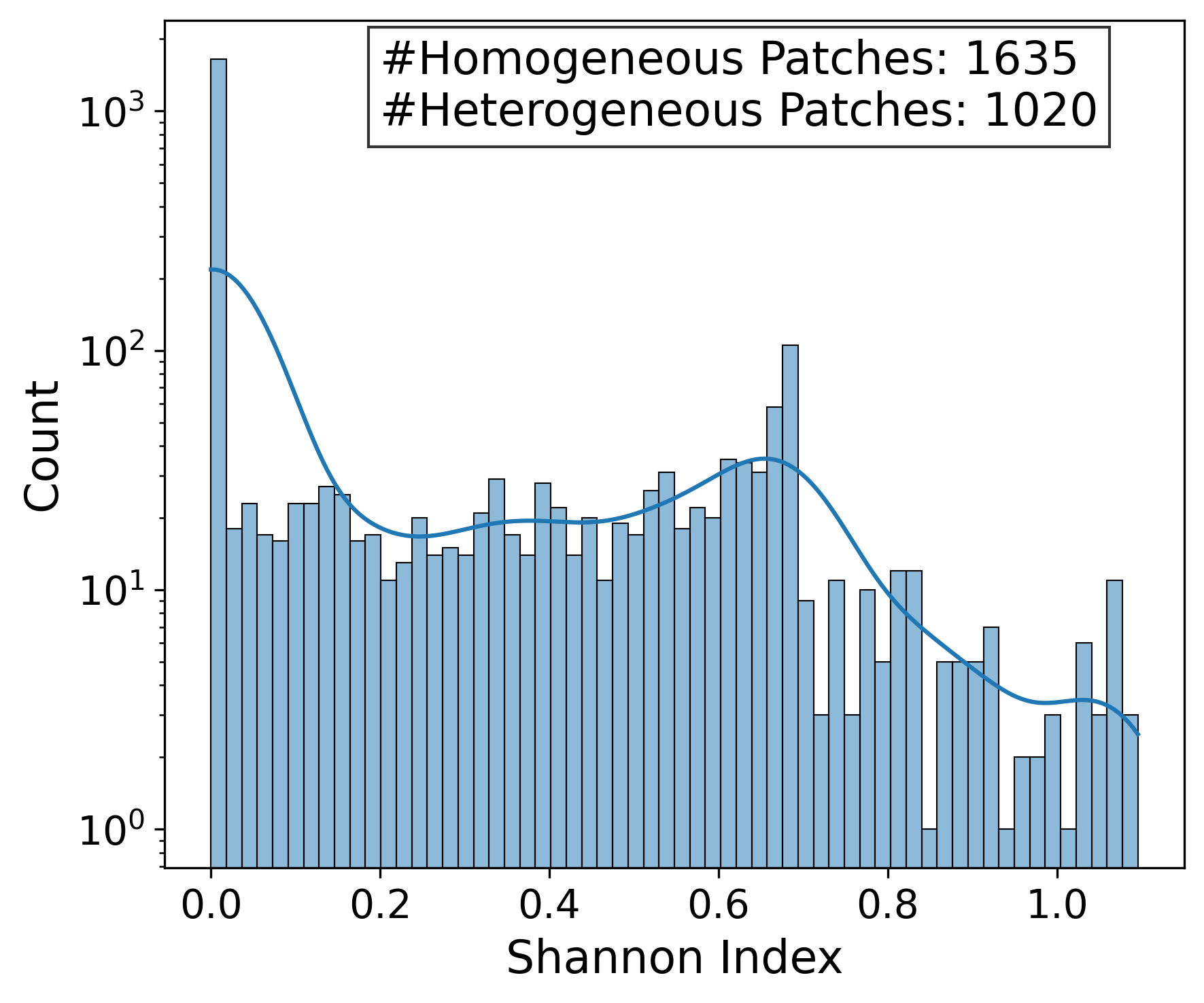}
        \caption{Arizona}
        \label{fig:az-shannon}
    \end{subfigure}
    \begin{subfigure}[b]{0.235\textwidth}
        \centering
        \includegraphics[width=\textwidth]{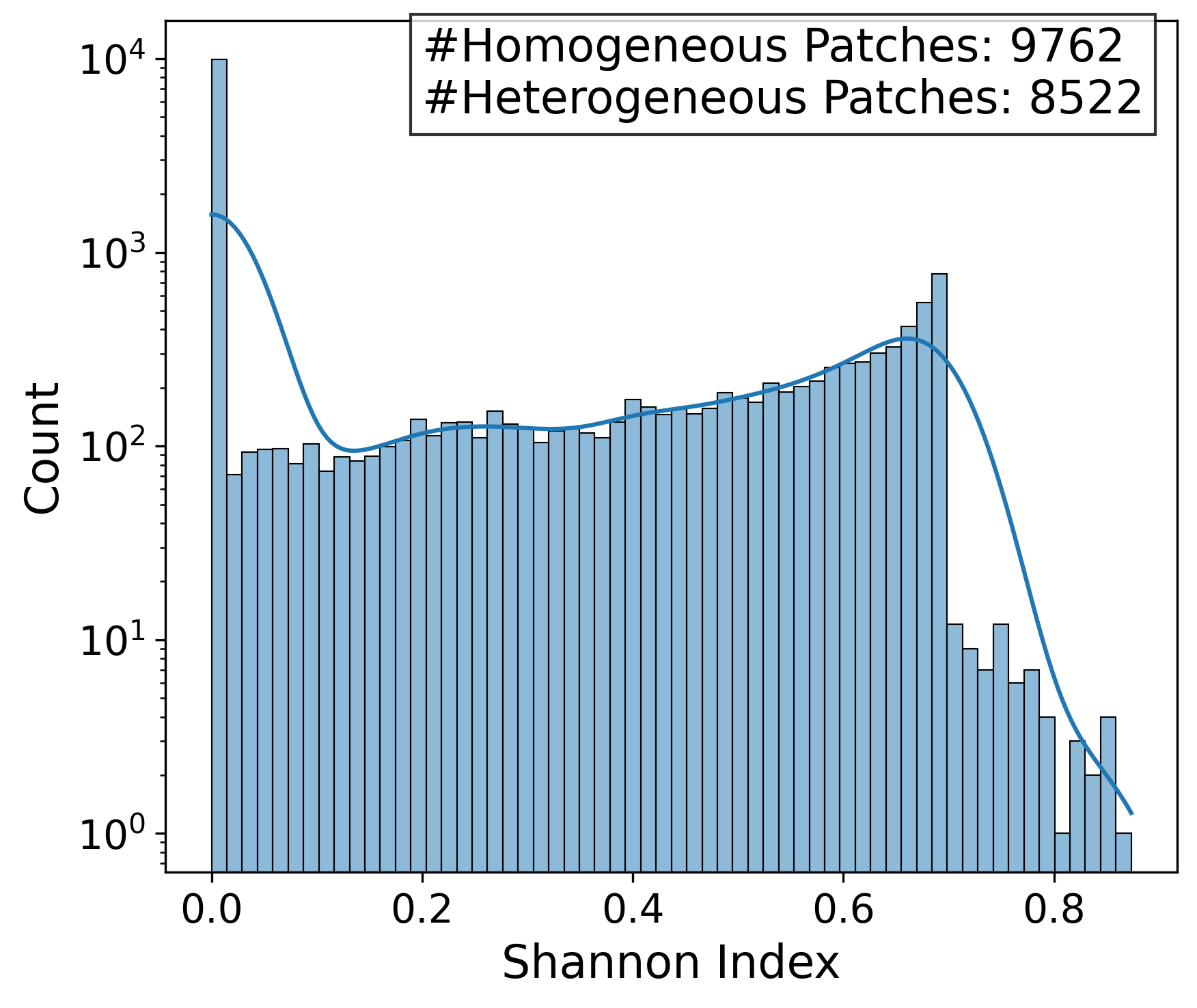}
        \caption{Colorado}
        \label{fig:co-shannon}
    \end{subfigure}
    \begin{subfigure}[b]{0.235\textwidth}
        \centering
        \includegraphics[width=\textwidth]{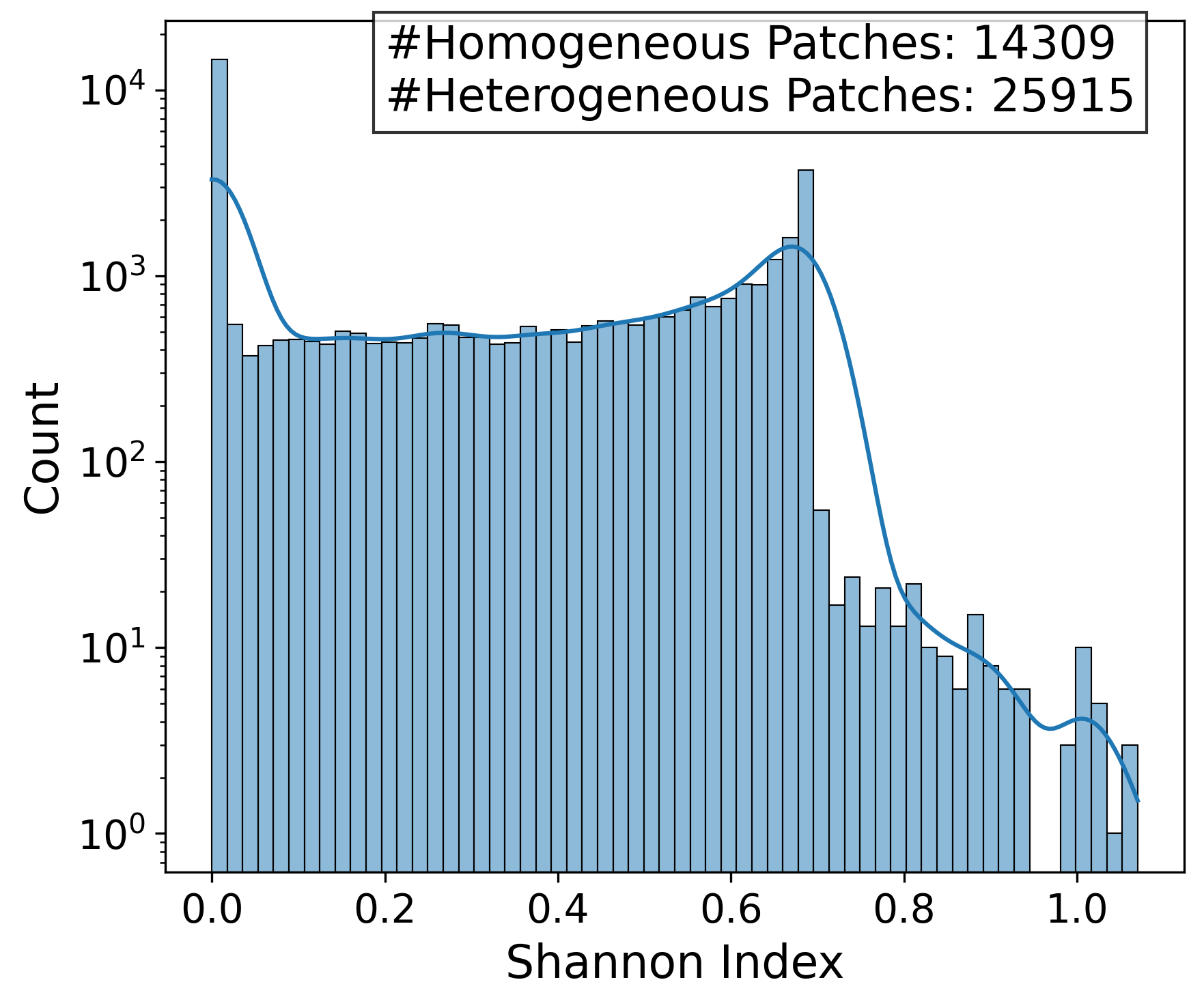}
        \caption{Utah}
        \label{fig:ut-shannon}
    \end{subfigure}
    \begin{subfigure}[b]{0.235\textwidth}
        \centering
        \includegraphics[width=\textwidth]{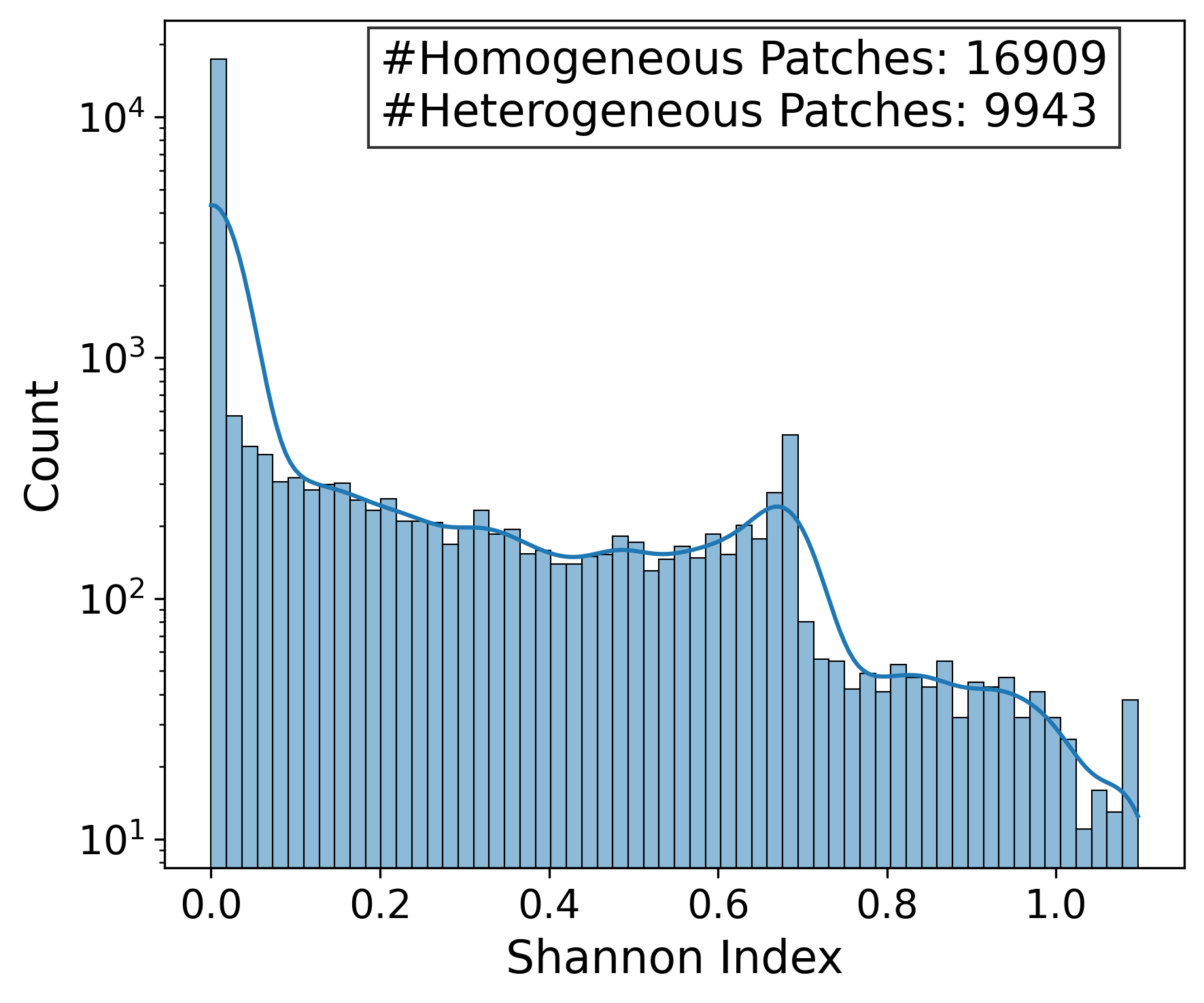}
        \caption{Washington}
        \label{fig:wa-shannon}
    \end{subfigure}
    
    \caption{The figure shows diversity of irrigation patterns across the states through Shannon diversity indices. Figure (a) presents Arizona's distribution; Figure (b) illustrates Colorado's similar pattern of irrigation concentration, with most patches having a Shannon index $= 0.0$; Figure (c) displays Utah's irrigation diversity, whereas Figure (d) shows Washington's irrigation diversity within patches.}
    \label{fig:shannon-diversity-state}
\end{figure*}

We analyze the diversity of the irrigation practices within the patches for each studied state. We notice that each state has a significant number of homogeneous patches (Shannon-index equal to 0) as shown in Figure \ref{fig:shannon-diversity-state}. However, for Utah and Colorado, the distribution of shannon-index  has a higher number of patches for the second peak (shannon-index equal to 0.7) compared to Arizona and Washington. Moreover, the numbers of heterogeneous patches are larger in Colorado and Utah compared to Arizona and Washington, indicating mixed irrigation practices in these two states.

\section{Properties of datasets}

\begin{table*}[!h]
\scriptsize
    \centering
    \caption{Summary of total storage size for LandSat and Sentinel per state and year. Sizes are given in GB or TB where appropriate.}
    \label{tab:data_summary}
    \renewcommand{\arraystretch}{1.2}
    \begin{tabular}{l|c|c|c}
        \toprule
        \textbf{State} & \textbf{Year} & \textbf{LandSat Size (GB/TB)} & \textbf{Sentinel Size (GB/TB)} \\
        \toprule
        AZ & 2013 & 78 GB &  \\
        \cline{2-4}
           & 2014 & 116 GB &  \\
        \cline{2-4}
           & 2015 & 94 GB & 42 GB \\
        \cline{2-4}
           & 2016 & 117 GB & 355 GB \\
        \cline{2-4}
           & 2017 & 133 GB & 447 GB \\
        \midrule
        CO & 2018 & 86 GB & 350 GB \\
        \cline{2-4}
           & 2019 & 65 GB & 352 GB \\
        \cline{2-4}
           & 2020 & 88 GB & 629 GB \\
        \midrule
        UT & 2021 & 72 GB & 501 GB \\
        \cline{2-4}
           & 2022 & 129 GB & 662 GB \\
        \cline{2-4}
           & 2023 & 117 GB & 435 GB \\
        \midrule
        WA & 2015 & 45 GB & 1 GB \\
        \cline{2-4}
           & 2016 & 53 GB & 83 GB \\
        \cline{2-4}
           & 2017 & 51 GB & 176 GB \\
        \cline{2-4}
           & 2018 & 45 GB & 266 GB \\
        \cline{2-4}
           & 2019 & 37 GB & 188 GB \\
        \cline{2-4}
           & 2020 & 47 GB & 312 GB \\
        \midrule
        \textbf{Total} &  & \textbf{1.4 TB} & \textbf{4.8 TB} \\
        \bottomrule
    \end{tabular}
\end{table*}
\begin{figure*}[h]
    \centering
    \begin{subfigure}[b]{0.4\textwidth}
        \centering
        \includegraphics[width=\textwidth]{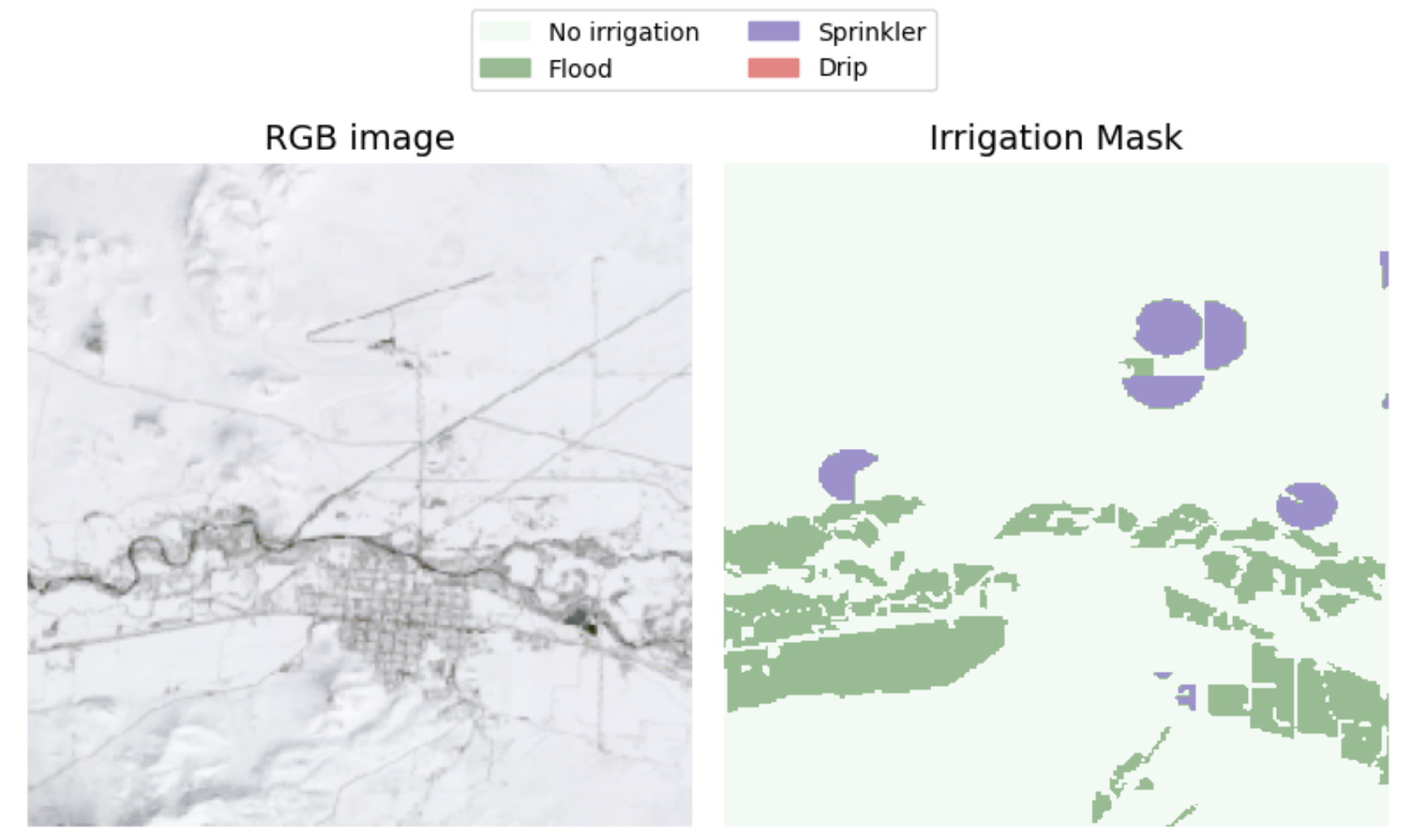}
        \caption{}
        \label{fig:snow-cover}
    \end{subfigure}
    \hspace{.2cm}
    \begin{subfigure}[b]{0.4\textwidth}
        \centering
        \includegraphics[width=\textwidth]{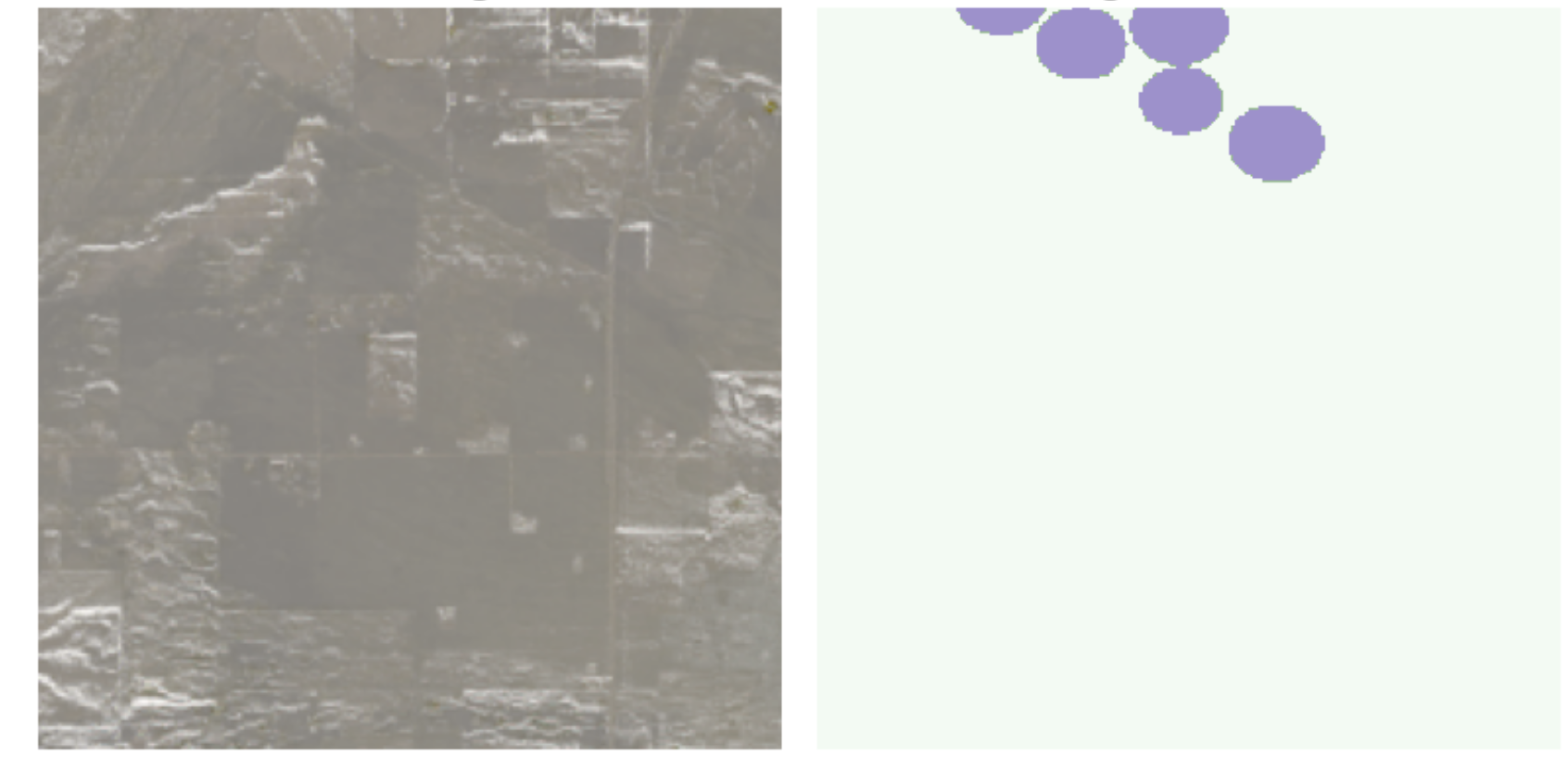}
        \caption{}
        \label{fig:bad-quality}
    \end{subfigure}
    \vskip\baselineskip
    \begin{subfigure}[b]{0.4\textwidth}
        \centering
        \includegraphics[width=\textwidth]{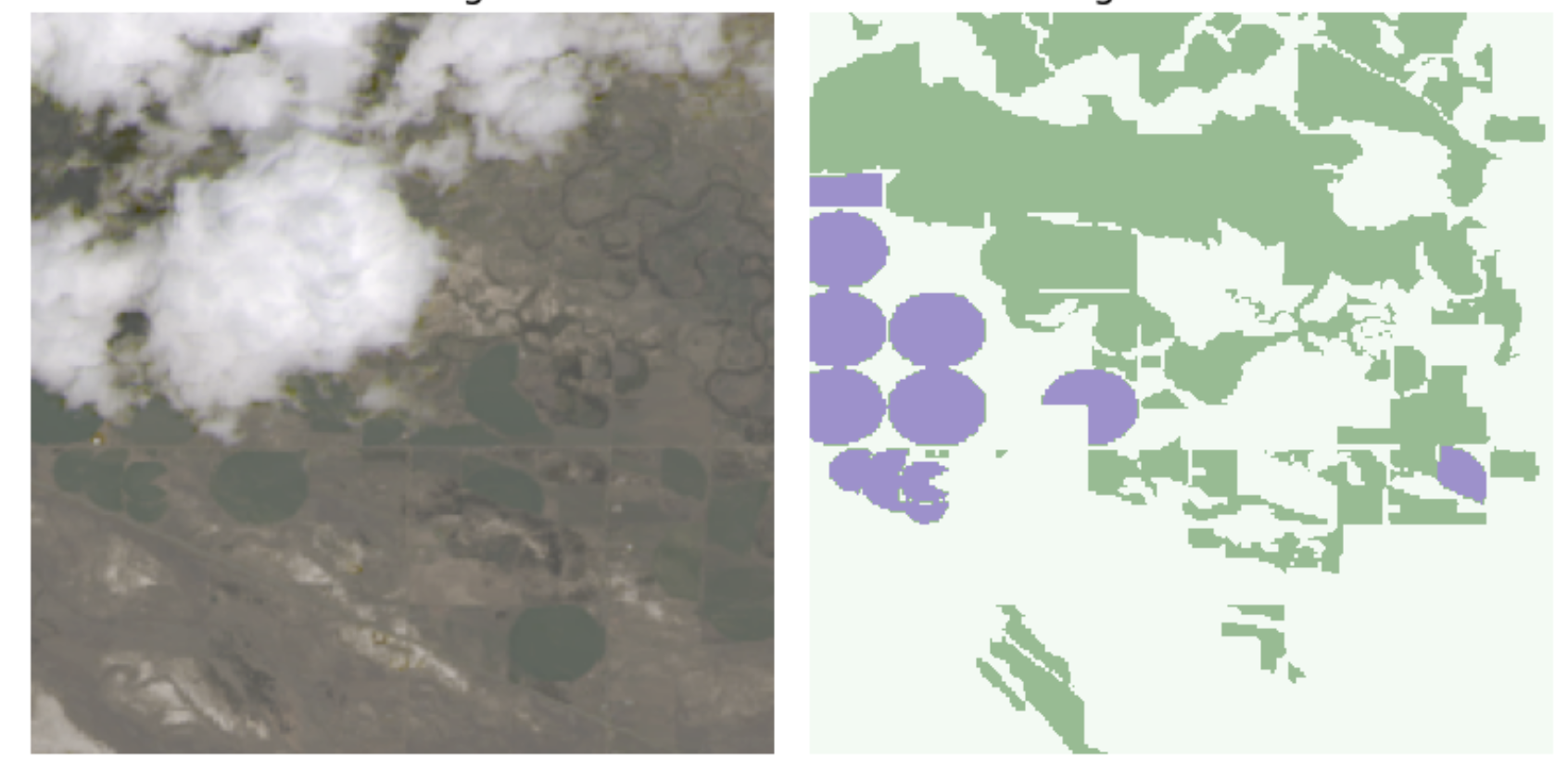}
        \caption{}
        \label{fig:cloud-cover}
    \end{subfigure}
    \hspace{.2cm}
    \begin{subfigure}[b]{0.4\textwidth}
        \centering
        \includegraphics[width=\textwidth]{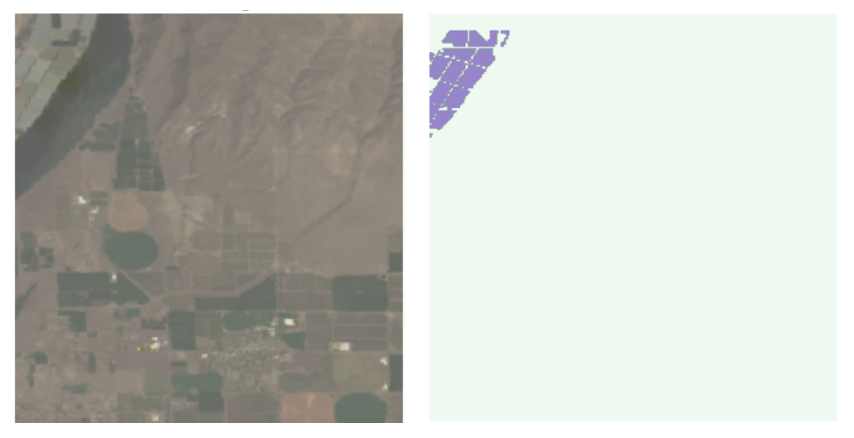}
        \caption{}
        \label{fig:incomplete}
    \end{subfigure}
    \caption{For each figure, the left image shows RGB images collected from satellite data, and the right figure shows the corresponding irrigation mask. (a) Agricultural land occluded by snow coverage, making irrigation methods identification difficult; (b) Poor image quality combined with snow coverage affecting landscape visibility; (c) Cloud and shadow occlusion impacting field visibility; and (d) Incomplete irrigation mask annotations with missing labels in the bottom region.}
    \label{fig:incomplete-images}
\end{figure*}

\begin{table*}[!ht]
\footnotesize
    \centering
    \renewcommand{\arraystretch}{3.5} 
    \caption{Summary of Common Vegetation Indices, Their Purpose, and Use Cases}
    \begin{tabular}{|p{3cm}|p{3.5cm}|p{3cm}|p{3cm}|}
        \hline
        \textbf{Index} & \textbf{Formula} & \textbf{Purpose} & \textbf{Common Use Cases} \\
        \hline
        NDVI (Normalized Difference Vegetation Index) & 
        $\frac{(NIR - Red)}{(NIR + Red)}$ & 
        Measures vegetation greenness and health & 
        Crop monitoring, land cover classification, drought detection \\
        \hline
        EVI (Enhanced Vegetation Index) & 
        $G \times \frac{(NIR - Red)}{(NIR + C_1 \times Red - C_2 \times Blue + L)}$ & 
        Reduces atmospheric and soil background effects & 
        More sensitive to high biomass regions \\
        \hline
        GNDVI (Green Normalized Difference Vegetation Index) & 
        $\frac{(NIR - Green)}{(NIR + Green)}$ & 
        More sensitive to chlorophyll content than NDVI & 
        Water stress detection, photosynthetic activity \\
        \hline
        SAVI (Soil-Adjusted Vegetation Index) & 
        $\frac{(NIR - Red)}{(NIR + Red + L)} \times (1 + L)$ & 
        Minimizes soil background influence & 
        Vegetation monitoring in arid or semi-arid areas \\
        \hline
        MSAVI (Modified Soil-Adjusted Vegetation Index) & 
        $\frac{2NIR + 1 - \sqrt{(2NIR + 1)^2 - 8(NIR - Red)}}{2}$ & 
        Further reduces soil influence compared to SAVI & 
        Useful for sparse vegetation and dry land monitoring \\
        \hline
        RVI (Ratio Vegetation Index) & 
        $\frac{NIR}{Red}$ & 
        Alternative to NDVI, less sensitive to atmospheric conditions & 
        Biomass and vegetation density analysis \\
        \hline
        CIgreen (Chlorophyll Index) & 
        $\frac{NIR}{Green} - 1$ & 
        Estimates chlorophyll content & 
        Plant health monitoring \\
        \hline
        NDWI (Normalized Difference Water Index) & 
        $\frac{(NIR - SWIR)}{(NIR + SWIR)}$ & 
        Measures water content in vegetation & 
        Drought monitoring, irrigation management \\
        \hline
        PRI (Photochemical Reflectance Index) & 
        $\frac{(Green - Blue)}{(Green + Blue)}$ & 
        Measures plant stress and efficiency & 
        Photosynthesis monitoring \\
        \hline
        OSAVI (Optimized Soil-Adjusted Vegetation Index) & 
        $\frac{(NIR - Red)}{(NIR + Red + 0.16)}$ & 
        An improved version of SAVI that minimizes soil background effects while maintaining sensitivity to vegetation. & 
        Used for vegetation monitoring in areas with moderate soil exposure. \\
        \hline
        WDRVI (Wide Dynamic Range Vegetation Index) & 
        $\frac{a \times NIR - Red}{a \times NIR + Red}$ & 
        A modified NDVI that enhances sensitivity to vegetation changes in high biomass areas. & 
        Used in precision agriculture to track crop growth and stress detection. \\
        \hline
        NDTI (Normalized Difference Tillage Index) & 
        $\frac{(SWIR1 - SWIR2)}{(SWIR1 + SWIR2)}$ & 
        Differentiates between tilled and untilled soil, helping in soil disturbance and land management analysis. & 
        Applied in soil erosion studies and land conservation planning. \\
\bottomrule
    \end{tabular}
    \label{tab:vegetation_indices}
\end{table*}

\begin{table*}
\small
\centering
\caption{Mapping of Individual Crops to Crop Groups. The `UNK' crop group indicates the crops can not be specified in any crop groups.}
\label{tab:crop_mapping}
\resizebox{\textwidth}{!}{
\begin{tabular}{ll}
\hline
\textbf{Crop Group} & \textbf{Individual Crops} \\
\hline
Alfalfa & Alfalfa, Alfalfa/Barley Mix, Alfalfa/Grass, New Alfalfa \\
\hline
Cereals & Barley, Barley/Wheat, Cereal Grain, Corn, Durum Wheat, Grain/Seeds unspecified, Oats, Rye, \\
& Sorghum, Speltz, Spring Wheat, Triticale, Wheat, Winter Wheat, Corn Grain, Corn Silage, \\
& Small Grains, Sorghum Grain, Spring Grain, Sweet Corn, Wheat Fall, Wheat Spring, Field Corn, \\
& Double crop barley/corn, Double crop winter wheat/corn \\
\hline
Cover Crop & Cover Crop, Green Manure, Field Crops, Other Field Crops \\
\hline
Fibres & Cotton \\
\hline
Fruits & Apples, Apricots, Berries, Berry, Cherries, Citrus, Dates, Fruit Trees, Grapes, Melon, \\
& Oranges, Peaches, Pomegranate, Citrus Groves, Fruit \\
\hline
Grass & Bermuda Grass, Grass, Grass Hay, Hay/Silage, Idle Pasture, Other Hay/Non Alfalfa, Pasture, \\
& Pecan/Grass, Sod, Turfgrass, Turfgrass Ag, Turfgrass Urban, Grass Pasture, Bluegrass, \\
& Sod Farm, Grass/Hay/Pasture, Hay, Improved Pasture - Irrigated, Rye Grass, Grassland/Pasture, \\
& Irrigated turf \\
\hline
Green House & Greenhouse \\
\hline
Herb Group & Flowers, Herb \\
\hline
Horticulture & Horticulture \\
\hline
Nursery & Nurseries, Nursery, Nursery Trees, Tree Nurseries, Tree Nursery \\
\hline
Nuts & Almond, Pecans, Pistachios, Walnuts \\
\hline
Oil-bearing crops & Canola, Flaxseed, Jojoba, Mustard, OilSeed, Olives, Safflower, Soybeans \\
\hline
Orchard & Orchard, Orchard unspecified, Orchard With Cover, Orchard Wo Cover \\
\hline
Pulses & Beans, Dry Beans, Garbanzo, Seed, Peanuts, Seeds \\
\hline
Roots and Tubers & Potato, Potatoes \\
\hline
Shrub Plants & Guayule, Shrubland \\
\hline
Sugar Crops & Sugar Beets, Sugarbeets, Sunflower, Sugar Cane, Sugar cane \\
\hline
UNK & Commercial Tree, Fallow, Fallow/Idle, Field Crop unspecified, Idle, Not Specified, Other, \\
& Sudan, Transition, Trees, Urban, Ornamentals, Research Facility, Research land, \\
& Miscellaneous vegetables and fruits, Other tree crops \\
\hline
Vegetables & Flower Bulb, Lettuce, Onion, Pumpkins, Squash, Vegetable, Vegetables, Watermelons, \\
& Eggplant, Fall Vegetables, Spring Vegetables, Vegetables Double Crop, Cabbage, Onions, Peppers \\
\hline
Vineyard & Vineyard \\
\hline
\end{tabular}
}
\end{table*}


\end{document}